\documentclass{article}

 \usepackage[preprint]{neurips_2026}

\usepackage[utf8]{inputenc} 
\usepackage[T1]{fontenc}    
\usepackage{hyperref}       
\usepackage{url}            
\usepackage{booktabs}       
\usepackage{amsfonts}       
\usepackage{nicefrac}       
\usepackage{microtype}      
\usepackage{xcolor}         
\usepackage{graphicx}
\usepackage{multirow}
\usepackage{makecell}
\usepackage{placeins}
\usepackage[most]{tcolorbox}
\usepackage{float}
\usepackage{tabularx}
\usepackage{longtable}
\usepackage{ragged2e}
\usepackage{listings}
\usepackage{amssymb}
\usepackage{textcomp}
\usepackage{wrapfig}
\tcbuselibrary{skins}

\graphicspath{{figures/}}

\newtcolorbox{findingbox}{
  colback=gray!8,
  colframe=gray!50,
  boxrule=0.5pt,
  arc=2pt,
  left=6pt,
  right=6pt,
  top=5pt,
  bottom=5pt,
  before skip=8pt,
  after skip=8pt
}

\title{CRAFT: Clinical Reward-Aligned Finetuning for Medical Image Synthesis}

\author{%
\textbf{Yunsung Chung}\textsuperscript{1} \quad
\textbf{Alex El Darzi}\textsuperscript{2} \quad
\textbf{Carlo El Khoury}\textsuperscript{2}
\\
\textbf{Han Feng}\textsuperscript{\textbf{2}} \quad
\textbf{Nassir Marrouche}\textsuperscript{\textbf{2}} \quad
\textbf{Jihun Hamm}\textsuperscript{\textbf{1}}
\\[4pt]
\textsuperscript{1}Department of Computer Science, Tulane University, New Orleans, LA 70118 \\
\textsuperscript{2}School of Medicine, Tulane University, New Orleans, LA 70112 \\
\texttt{\{ychung3,jhamm3\}@tulane.edu}
}

\begin{document}

\maketitle

\begin{abstract}
Foundation diffusion models can generate photorealistic natural images, but adapting them to medical imaging remains challenging.
In medical adaptation, limited labeled data can exacerbate hallucination-like and clinically implausible synthesis, while existing metrics such as FID or Inception Score do not quantify per-image alignment with pathology-relevant criteria.
We introduce the Clinical Alignment Score (CAS), a foundation-model-based proxy for clinical alignment that evaluates generated images along four complementary dimensions beyond visual fidelity.
Building on CAS, we propose Clinical Reward-Aligned Finetuning (CRAFT), a reward-based adaptation framework that transfers medical knowledge from multimodal large language models and vision-language models through label-conditioned prompt enrichment, clinical checklists, and differentiable reward optimization.
Across four diverse modalities, CRAFT improves CAS and downstream classification performance over strong adaptation baselines.
Beyond average CAS gains, CRAFT reduces the empirical low-alignment tail below a real-image reference threshold by 5.5--34.7\% points relative to the strongest baseline, corresponding to a 20.4\% average relative reduction across datasets.
These results indicate fewer hallucination-like generations under CAS, and are corroborated by out-of-family evaluator evaluation, structured checklist auditing, memorization analysis, and a blinded physician preference study on CheXpert.
\end{abstract}

\section{Introduction}
\label{sec:intro}

\begin{wrapfigure}{r}{0.45\textwidth}
  \vspace{-10pt}
  \centering
  \includegraphics[width=\linewidth]{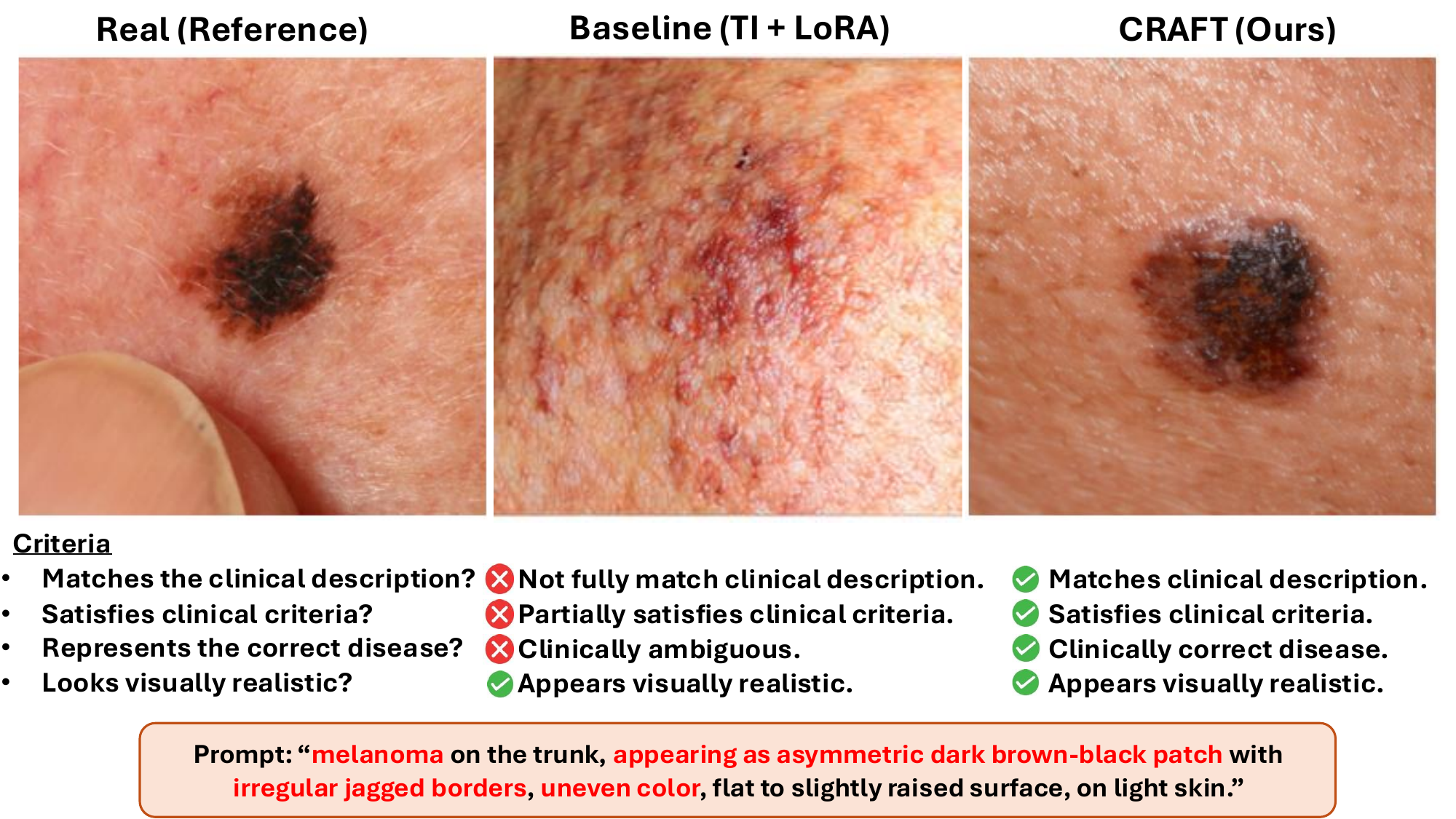}
  \caption{
  Melanoma example. The baseline synthesis lacks clear pathology-specific cues, while CRAFT exhibits stronger alignment with the clinical criteria captured by CAS.
  }
  \label{fig:first}
  \vspace{-10pt}
\end{wrapfigure}

Deep learning has demonstrated remarkable success in medical decision-making, including dermatology and radiology \citep{anderson2024deep,esteva2017dermatologist, brinker2019convolutional, liu2020deep,soenksen2021using}. However, robust medical model training remains constrained by limited labeled data, privacy barriers, annotation cost, and long-tailed clinical distributions.

Synthetic data generation has emerged as a partial remedy to medical data scarcity \citep{chung2025sok}. 
Diffusion models \citep{ho2020denoising} have been adapted from general \citep{gal2022image, kim2022diffusionclip, li2023blip, ruiz2023dreambooth, ye2023ip, zhang2023adding} to medical domains, including chest X-ray, brain MRI, CT, and dermatology \citep{ali2022spot,huang2024chest,khader2023denoising,pinaya2022brain,guo2025maisi,xu2024medsyn,wang2025doctor}. 
Synthetic data can also improve downstream utility or mitigate distribution shifts in some settings \citep{chung2024bridging,wang2024majority,ktena2024generative}.

In medical imaging, visual realism is insufficient: a generated image can look plausible while missing the pathology-specific findings that make it clinically useful.
Distribution-level metrics such as Fréchet Inception Distance (FID) \citep{heusel2017gans}, Inception Score (IS) \citep{salimans2016improved}, and Kernel Inception Distance \citep{binkowski2018demystifying} measure set-level visual fidelity, while LPIPS \citep{zhang2018unreasonable} and CLIPScore \citep{hessel2021clipscore} measure perceptual or text-image alignment. None directly evaluates whether a generated image satisfies pathology-relevant criteria for a given clinical prompt (see Figure~\ref{fig:first}).

To close this gap, we propose the Clinical Alignment Score (CAS), a foundation-model-based proxy for clinical alignment. CAS combines four primitives: Visual Description Consistency (VDC), Clinical Criteria Satisfaction (CCS), Diagnostic Discriminability (DD), and Semantic Feature Similarity (SFS). 
Unlike generic perceptual metrics, CAS targets clinically meaningful failure modes via text–image alignment, checklist satisfaction, diagnostic separability, and reference-anchored semantic similarity, and it requires no manual preference labels. CAS remains a proxy, not a substitute for expert review.

Building on CAS, we introduce CRAFT (Clinical Reward-Aligned Finetuning), which turns the same clinical dimensions into differentiable training signals. 
CRAFT adapts a general-purpose diffusion model toward clinically aligned medical synthesis under limited supervision. 
Specifically, we use MLLMs (e.g., MedGemma \citep{sellergren2025medgemma}, GPT-4o \citep{achiam2023gpt}) to expand sparse clinical labels into image-grounded descriptions and create disease-level clinical criteria, and also use VLMs (e.g., MedSigLIP \citep{sellergren2025medgemma}) as a differentiable critic that supplies reward feedback during finetuning.
Across four medical imaging datasets, CRAFT achieves higher CAS on held-out test cases than the compared adaptation baselines, and also improves real+synthetic downstream classification performance.
On CheXpert, CRAFT samples are rated higher by medical experts, supporting improved clinical plausibility beyond simple memorization.
While we focus on medical imaging, the framework should extend naturally to other domains with expert-defined semantic criteria---e.g., satellite imagery, materials science, or industrial inspection.

Clinical alignment must be evaluated not only by average performance but also by failure tails.
In medical image synthesis, even a small fraction of hallucination-like samples can inject misleading pathology cues into downstream training. 
We therefore define a low-alignment rate using a real-image reference threshold and analyze the lower tail of per-image CAS. Across four datasets, CRAFT reduces this low-alignment rate by 5.5--34.7\% points relative to the strongest baseline, indicating that clinical reward alignment improves not only average quality but also tail reliability.

Our contributions are threefold.
(1) We introduce CAS, a decomposed foundation-model-based proxy for clinical alignment that combines VDC, CCS, DD, and SFS.
(2) We propose CRAFT, a reward-aligned finetuning framework that transfers structured medical knowledge from MLLMs and VLMs into a diffusion model through prompt enrichment and differentiable reward optimization. 
(3) Across dermatology, radiology, histopathology, and retinal fundus imaging, CRAFT improves average CAS, reduces low-CAS failure rates, and improves downstream classification accuracy. We corroborate these gains with out-of-family evaluator analysis, checklist auditing, diversity analysis, and blinded physician preference study on CheXpert.



\begin{figure*}[!t]
  \vskip 0.2in
  \begin{center}
    \centerline{\includegraphics[width=0.85\linewidth]{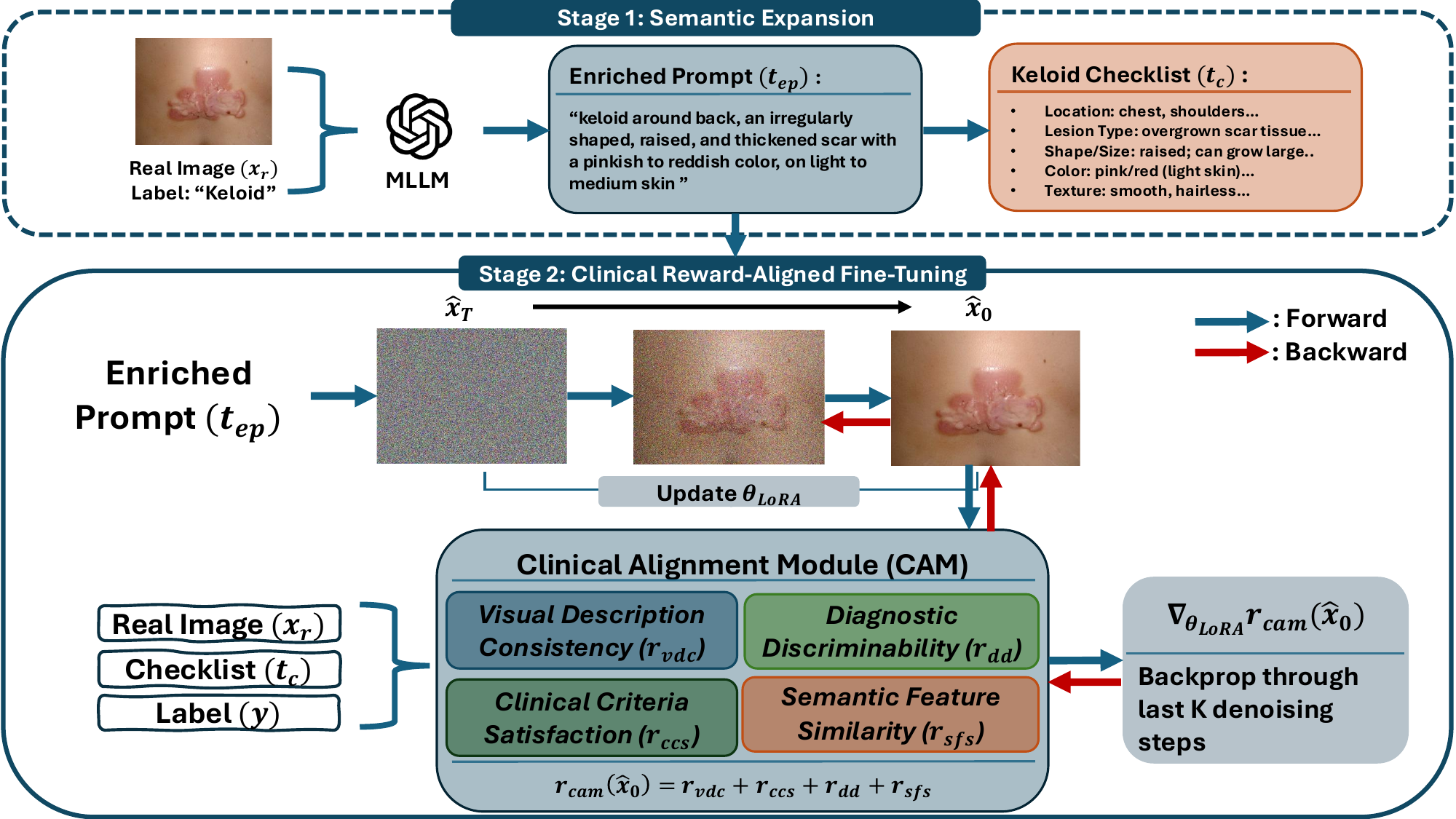}}
    \caption{
      Overview of the CRAFT Framework. The pipeline consists of two stages: (1) Semantic Enrichment uses an MLLM to generate an image-grounded enriched prompt ($t_{ep}$) from each image-label pair ($x_i,y_i$) and a class-level checklist ($t_c$) from the label $y_i$. (2) Clinical Reward-Aligned Finetuning optimizes a pretrained diffusion model using a VLM critic, which computes differentiable rewards for visual description consistency ($r_\textrm{vdc}$), clinical criteria satisfaction ($r_\textrm{ccs}$), diagnostic discriminability ($r_\textrm{dd}$), and semantic feature similarity ($r_\textrm{sfs}$). Reward gradients are backpropagated through the final $K$ denoising steps, updating only LoRA parameters for efficient training.
    }
    \label{fig:framework}
  \end{center}
  \vspace{-5pt}
\end{figure*}

\section{Related works}
\textbf{Medical image synthesis.}
Latent diffusion models \citep{rombach2022high} have been adapted to medical imaging domains such as chest X-rays \citep{ali2022spot, de2023medical, huang2024chest}, 3D brain MRI \citep{khader2023denoising, pinaya2022brain}, and CT \citep{guo2025maisi, zhao2025maisi}. Beyond high-fidelity, synthetic data can improve robustness in data-scarce regimes \citep{wang2025doctor}, and mitigate demographic distribution shifts \citep{ktena2024generative, wang2024majority}. To introduce semantic control, prior work adapts pretrained diffusion models using Textual Inversion \citep{gal2022image} and/or LoRA \citep{hu2022lora} to inject disease concepts into pre-trained models \citep{de2023medical,wang2024majority,fayyad2025lesiongen}. However, optimizing for distributional realism or reconstruction fidelity alone does not guarantee pathology-specific correctness. Prior work has shown that medical image translation and synthesis systems can introduce hallucinated features when objectives fail to preserve clinically meaningful structure \citep{cohen2018distribution}.

\textbf{Reward and preference alignment.}
Diffusion model alignment increasingly uses reward or preference signals, extending ideas such as DPO \citep{rafailov2023direct} to images via offline preference learning and reward modeling \citep{wallace2024diffusion, xu2023imagereward}. 
Existing methods include classifier/reward-model-based alignment \citep{bai2025towards, li2024aligning, li2025divergence, wallace2024diffusion} and direct optimization through the denoising process \citep{black2023training, clark2023directly}. 
In medical imaging, \citet{sun2023aligning} leverages direct pathologist feedback, while subsequent work explores automated feedback: \citet{wang2025doctor} utilizes DPO to align skin lesion synthesis, and \citet{saremi2025rl4med} optimizes against online classifiers but relies on discrete reward proxies. 
CRAFT differs in jointly combining label-conditioned prompt enrichment, class-level clinical checklists, reference-anchored semantic rewards, and differentiable reward optimization within a unified framework, and in evaluating across four imaging domains rather than one.

\section{Methods}
Diffusion models pretrained on general-domain images provide strong priors over natural image statistics but lack the domain-specific knowledge needed to interpret sparse clinical labels.
To bridge this gap, we propose CRAFT (Clinical Reward-Aligned Finetuning), which adapts a general-purpose diffusion model toward clinically aligned medical image synthesis through two stages (Figure~\ref{fig:framework}).



\subsection{Stage 1: LLM-Driven semantic enrichment}
Let $D=\{(x_i,y_i)\}$ be a dataset of medical images with sparse clinical labels $y_i\in Y$. Conditioned on $y_i$ alone, a pretrained diffusion model often produces generic or clinically inconsistent samples, reflecting the limited coverage of medical concepts in its pretraining data. We address this with MLLM-generated descriptions that ground each sparse label in the paired training image. For each image-label pair $(x_i,y_i)$, we generate two complementary textual artifacts:

\textbf{Enriched Prompt ($t_{ep}$).} A per-image visual description generated by the MLLM that captures instance-specific visual attributes of $x_i$, such as morphology, color variation, and border characteristics, in a domain-dependent manner.
The ground-truth label $y_i$ is provided as a constraint during prompt generation to suppress diagnostic hallucination and enforce consistency with the underlying pathology.
This enriched prompt translates an abstract label into image-grounded visual cues understandable by non-expert models (e.g., ``an asymmetric hyperpigmented lesion with irregular jagged borders''), enabling the diffusion model to condition on fine-grained, instance-level appearance (Appendix \ref{app:prompt}).

\textbf{Natural Language Checklist ($t_c$).} A per-disease structured checklist that encodes label-level clinical criteria shared across all images of class $y$ (e.g., required shape, texture, and color).
Unlike $t_{ep}$, which varies across images, $t_c$ is fixed for each disease and serves as a stable clinical constraint that enforces clinical consistency across generations (Appendix \ref{app:checklists}).

\subsection{Stage 2: clinical reward-aligned finetuning}
In the second stage, we finetune the diffusion model parameters $\theta$ by combining the standard diffusion objective with a clinical reward computed by a frozen VLM critic. 
Because Stable Diffusion is a latent diffusion model, the diffusion reconstruction term is computed in VAE latent space \citep{kingma2013auto}, while clinical rewards are computed after decoding the generated latent into pixel space.

Let $z_r=\mathcal{E}_{\rm VAE}(x_r)$ denote the VAE latent of a real training image $x_r$, and let $z_t=\alpha_t z_r+\sigma_t\epsilon$ be the noised latent at diffusion timestep $t$. 
The text condition is the enriched prompt $c=t_{ep}$. 
During reward optimization, differentiable sampling through the final $K$ denoising steps produces a generated latent $\hat{z}_0$, which is decoded into an image $\hat{x}_0=\mathcal{D}_{\rm VAE}(\hat{z}_0)$ before computing VLM-based rewards. 
The training objective is:
\begin{equation}
\mathcal{L}(\theta)=
\mathbb{E}_{z_r,c,t,\epsilon}
\left[
\lambda_{\rm diff}\|\epsilon-\epsilon_\theta(z_t,t,c)\|_2^2
-
\lambda_{\rm cam}r_{\rm cam}(\hat{x}_0,x_r,c,t_c,y)
\right].
\end{equation}

The reference image $x_r$ is used for the diffusion reconstruction term and SFS reward during training, but not provided to the generator at test-time synthesis.
To preserve the model's generative priors, we employ LoRA to optimize only low-rank decomposition matrices added to the attention layers. Gradients from $r_\textrm{cam}$ are propagated through the sampled image $\hat{x}_0$ and the final $K$ denoising steps while the VLM encoders remain fully frozen, similar to DRaFT-$K$ \citep{clark2023directly}.

The reward $r_{\rm cam}$ is computed by the Clinical Alignment Module (CAM) as a sum of four components:
\textbf{Objective 1: Visual Description Consistency (VDC) Rewards.}
To ensure that the synthesized image reflects the fine-grained visual attributes described in the enriched prompt generated in Stage~1, we maximize the cosine similarity between the image embedding $E_I(\hat{x}_0)$ and the text embedding of the image-grounded enriched prompt $t_{ep}$:
\begin{equation}
    r_\textrm{vdc} = \cos\big(E_I(\hat{x}_0), E_T(t_{ep})\big)
\end{equation} 
This reward enforces alignment with \emph{instance-specific} morphology, color variation, and structural details, preserving fine-grained semantic fidelity while leaving room for visual diversity across samples.

\textbf{Objective 2: Clinical Criteria Satisfaction (CCS) Rewards.}
To ensure adherence to disease-specific clinical criteria shared across samples of the same label, we additionally enforce alignment with structured clinical criteria expressed as a natural language checklist $t_c$:

\begin{equation}
    r_\textrm{ccs} = \cos\big(E_I(\hat{x}_0), E_T(t_{c})\big)
\end{equation}
Unlike $r_\textrm{vdc}$, which captures instance-level appearance, $r_\textrm{ccs}$ enforces \emph{class-level} clinical consistency, such as required lesion morphology or anatomical patterns. This separation enables the model to generate visually diverse samples while encouraging alignment with disease-specific visual criteria.

\textbf{Objective 3: Diagnostic Discriminability (DD) Reward.}
Visual similarity to text does not guarantee diagnostic separability. To ensure the generated features are discriminative for the correct pathology $y$, we employ a linear probe $\phi$ trained on the frozen VLM embedding space using the same training split used to finetune the diffusion model. We define the reward as the log-likelihood of the correct class:
\begin{equation} 
    r_\textrm{dd} = \log \frac{\exp(\phi(E_I(\hat{x}_0))_y)}{\sum_{k=1}^C \exp(\phi(E_I(\hat{x}_0))_k)}
\end{equation}
Maximizing this reward penalizes generations that look realistic but lack the feature combinations required for confident diagnosis. 

\textbf{Objective 4: Semantic Feature Similarity (SFS) Reward.}
A common failure mode in synthetic medical data is hallucinatory drift: models produce exaggerated depictions that lack the fine-grained textural fidelity of real pathology \citep{cohen2018distribution}. To discourage such drift, we introduce a reference-anchored semantic similarity objective that operates in a frozen medical VLM embedding space. For each training step, the generated sample $\hat{x}_0$ is paired with its corresponding real reference image $x_r$:
\begin{equation}
r_\textrm{sfs}=\cos\Big(E_I(\hat{x}_0),E_I(x_r)\Big)
\end{equation}
$r_\textrm{sfs}$ anchors generations to real medical-image representation and discourages semantic drift toward exaggerated or unrealistic pathology.
It is used during paired training and reference-based evaluation, but not for prompt-only deployment.
Appendix~\ref{app:memorization} finds no pixel-level or feature-space memorization.

\subsection{Optimization via DRaFT}
We optimize $r_{\rm cam}$ using DRaFT-$K$ \citep{clark2023directly}, which backpropagates reward gradients only through the final $K$ denoising steps, reducing memory cost and instability. We update only LoRA parameters and keep the VLM encoders frozen. To reduce gradient variance from stochastic sampling, we generate $M$ trajectories per prompt and average rewards:



\begin{equation}
r_{\rm cam}
=
\frac{1}{M}\sum_{m=1}^{M}
\frac{1}{4}
\left(
r_{\rm vdc}^{(m)}+
r_{\rm ccs}^{(m)}+
r_{\rm dd}^{(m)}+
r_{\rm sfs}^{(m)}
\right).
\end{equation}
We use $\lambda_{\rm diff}=0.2$ and $\lambda_{\rm cam}=0.8$, giving each reward component an effective weight of $0.2$ in the full objective. Appendix~\ref{app:dd_weight} reports DD-weight sensitivity.

\section{Experiments}
\subsection{Experimental setup}
\subsubsection{Datasets}
We evaluate CRAFT on four medical imaging domains with distinct visual characteristics.

\textbf{Fitzpatrick17k.} We use the 20-condition subset of Fitzpatrick17k \citep{groh2021evaluating} following \citep{wang2025doctor}, with a 50/50 patient-level split (3,100 train / 3,100 test).

\textbf{CheXpert.} Following \citep{chung2025sok}, we use a 4-class frontal-view CheXpert subset \citep{irvin2019chexpert} (\textit{Cardiomegaly}, \textit{Pleural Effusion}, \textit{Pneumonia}, \textit{No Finding}), totaling 990 images (495 train / 495 test).

\textbf{BreakHis.} We use the 100$\times$ BreakHis subset \citep{xie2019deep} with 8 diagnostic categories and a 50/50 split (1,040 train / 1,041 test).

\textbf{ORIGA.} We use the ORIGA retinal fundus dataset \citep{zhang2010origa} with glaucoma / non-glaucoma labels, following the official split (454 train / 196 evaluation).


\subsubsection{Baselines}
We compare against SD2.1 zero-shot, Textual Inversion (TI) \citep{de2023medical}, TI+LoRA \citep{wang2024majority}, and DPO-style preference optimization \citep{wang2025doctor}. Following \citet{wang2025doctor}, DPO uses 1,024 offline preference pairs per dataset, with MedGemma selecting the preferred image from two seed-varied LoRA samples per pair. We include zero-shot Janus-Pro and Flux as general-purpose references in Appendix~\ref{app:zeroshot_generators}.

\begin{table*}[t]
\centering
\small
\setlength{\tabcolsep}{2pt}
\caption{Quantitative results across four medical imaging domains. Synthetic-method values are means over five seeds. The ``Real'' row is a reference point, not a target (see Sec.~\ref{sec:quantitative}). CRAFT achieves the best CAS on all four datasets.}
\resizebox{\linewidth}{!}{
\begin{tabular}{l|ccccc|ccccc|ccccc|ccccc}
\toprule
& \multicolumn{5}{c|}{\textbf{Fitzpatrick17k}}
& \multicolumn{5}{c|}{\textbf{CheXpert}}
& \multicolumn{5}{c|}{\textbf{BreakHis}}
& \multicolumn{5}{c}{\textbf{ORIGA}} \\
\cmidrule(lr){2-6} \cmidrule(lr){7-11} \cmidrule(lr){12-16} \cmidrule(lr){17-21}
Method
& VDC & CCS & DD & SFS & CAS
& VDC & CCS & DD & SFS & CAS
& VDC & CCS & DD & SFS & CAS
& VDC & CCS & DD & SFS & CAS \\
\midrule
Real
& .168 & .135 & .666 & 1.00 & .492
& .149 & .148 & .511 & 1.00 & .452
& .154 & .135 & .753 & 1.00 & .511
& .132 & .171 & .781 & 1.00 & .521 \\
\midrule
Zero-Shot
& .072 & .059 & .125 & .550 & .202
& .051 & .036 & .258 & .477 & .205
& .053 & .079 & .368 & .676 & .294
& .063 & .083 & .648 & .658 & .363 \\
DPO
& .093 & .121 & .350 & .760 & .331
& .060 & .119 & .252 & .848 & .320
& .121 & .124 & .413 & .852 & .378
& .080 & .156 & .734 & .878 & .462 \\
TI
& .092 & .119 & .339 & .757 & .327
& .060 & .119 & .255 & .848 & .320
& .128 & .104 & .418 & .820 & .367
& .127 & .153 & .697 & .847 & .456 \\
TI+LoRA
& .094 & .150 & .449 & .809 & .376
& .057 & .142 & .276 & \textbf{.941} & .354
& .142 & .120 & .477 & .872 & .402
& .137 & .170 & .800 & .880 & .497 \\
\midrule
\textbf{CRAFT}
& \textbf{.166} & \textbf{.158} & \textbf{.520} & \textbf{.824} & \textbf{.417}
& \textbf{.145} & \textbf{.146} & \textbf{.375} & .939 & \textbf{.401}
& \textbf{.153} & \textbf{.136} & \textbf{.486} & \textbf{.894} & \textbf{.417}
& \textbf{.147} & \textbf{.177} & \textbf{.842} & \textbf{.895} & \textbf{.515} \\
\bottomrule
\end{tabular}
}
\label{tab:combined_results}
\vspace{-12pt}
\end{table*}

\subsubsection{Implementation details}
All generators use Stable Diffusion v2.1 at 512$\times$512 resolution. We first learn disease-specific textual inversion tokens for 3,000 steps, then apply LoRA to UNet cross-attention layers and train for 10 epochs. CRAFT uses DRaFT-$K$ with $K=1$, $M=2$ stochastic repeats, and batch size 4. Additional training details are provided in Appendix~\ref{app:reproducibility}.


\textbf{Prompting and checklists.} Enriched prompts ($t_{ep}$) and class-level checklists ($t_c$) are generated offline from paired image-label data and then held fixed during optimization. Prompting details (\ref{app:prompt}) and templates (\ref{app:checklists}) are in the Appendix.

\textbf{Evaluation protocol.} Our main CAS evaluation follows a prompt-conditioned re-synthesis protocol: each held-out image is first converted into a label-constrained textual description, and the same text prompt is provided to every generator. The image pixels are never used as generator input and are used only after synthesis for reference-based metric computation such as SFS.

\subsubsection{Evaluation metrics}
To reduce evaluator leakage, CRAFT uses MedSigLIP as the frozen training critic, while all main quantitative metrics are computed with a separate SigLIP encoder not used for optimization. We further test robustness with MetaCLIP2 in Sec.~\ref{sec:eval_robustness}; model-role separation is summarized in Table~\ref{tab:role_separation}.

\textbf{VDC/CCS/SFS.} VDC and CCS are image--text cosine similarities between the generated image and the enriched prompt $t_{ep}$ or class-level checklist $t_c$, respectively. SFS is image--image cosine similarity between the generated image and its paired real reference in the evaluator embedding space.

\textbf{DD.} For training, the DD reward uses a linear probe trained on frozen MedSigLIP embeddings from real training images. For evaluation, we train a separate probe on frozen SigLIP embeddings and report classification accuracy on generated samples.

\textbf{CAS.} CAS is the macro-average of the four primitive scores (VDC, CCS, DD, SFS) on the held-out test split, with each primitive computed in the SigLIP evaluator's embedding space. Since CAS is a foundation-model-based proxy rather than clinical ground truth, we report primitive components and complement CAS with MetaCLIP2 evaluation, diversity analysis, downstream augmentation, checklist auditing, low-alignment tail analysis, and physician preference evaluation.


\begin{figure}[t]
  \centering
  \includegraphics[width=0.9\linewidth]{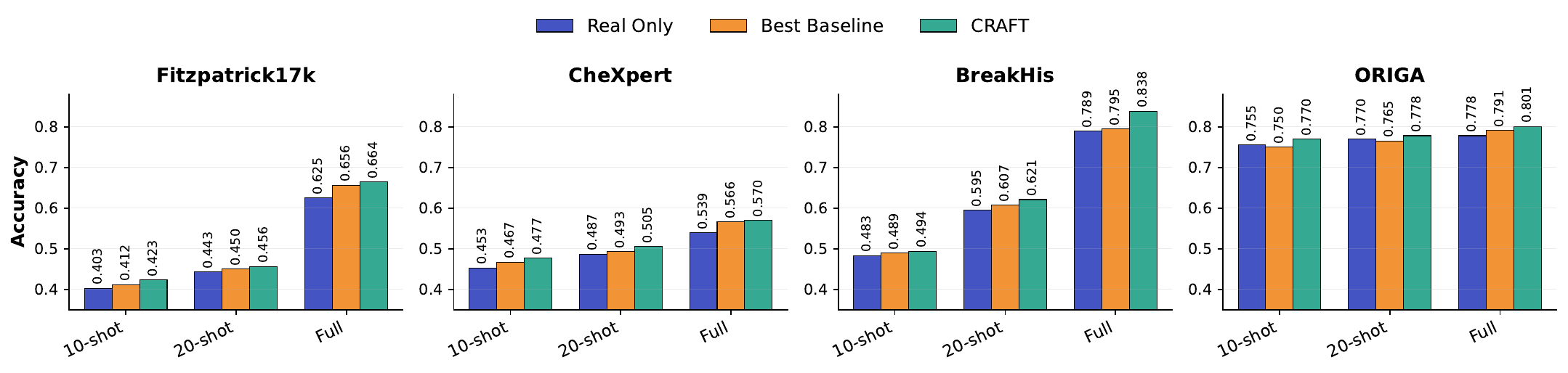}
  \caption{Downstream classification accuracy in a real+synthetic augmentation setting (20\% synthetic per batch) across four datasets and three data scales. CRAFT achieves the best accuracy in every setting. ``Best Baseline'' refers to the strongest non-reward method (TI+LoRA) per setting.}
  \label{fig:downstream_aug}
  \vspace{-5pt}
\end{figure}

\subsection{Main results}
We evaluate CRAFT on four distinct medical domains: dermatology (Fitzpatrick17k), radiology (CheXpert), histopathology (BreakHis), and retinal fundus imaging (ORIGA). We report CAS as the primary proxy for clinical alignment, and use downstream augmentation, out-of-family evaluation, checklist auditing, preference evaluation, and ablations as complementary evidence.

\subsubsection{Quantitative analysis}
\label{sec:quantitative}
Table~\ref{tab:combined_results} shows that CRAFT achieves the highest CAS among compared adaptation methods across all four domains. Relative to TI+LoRA, CRAFT improves CAS from 0.376 to 0.417 on Fitzpatrick17k, 0.354 to 0.401 on CheXpert, 0.402 to 0.417 on BreakHis, and 0.497 to 0.515 on ORIGA. The largest gains occur on Fitzpatrick17k and CheXpert, where clinically relevant morphology and subtle radiographic findings are difficult to convey from sparse labels alone. On Fitzpatrick17k, CRAFT improves diagnostic discriminability over TI+LoRA (0.520 vs.\ 0.449), while on CheXpert, it substantially improves VDC (0.145 vs.\ 0.057), indicating stronger evaluator-measured alignment between generated images and clinical text conditions.

The ``Real'' row is not a theoretical upper bound on every component: VDC and CCS depend on text--image embedding alignment and checklist phrasing, while SFS is 1.0 by construction (each real image is compared to itself).

\subsubsection{Low-alignment tail analysis}
\begin{figure*}[t]
  \centering
  \includegraphics[width=0.95\linewidth]{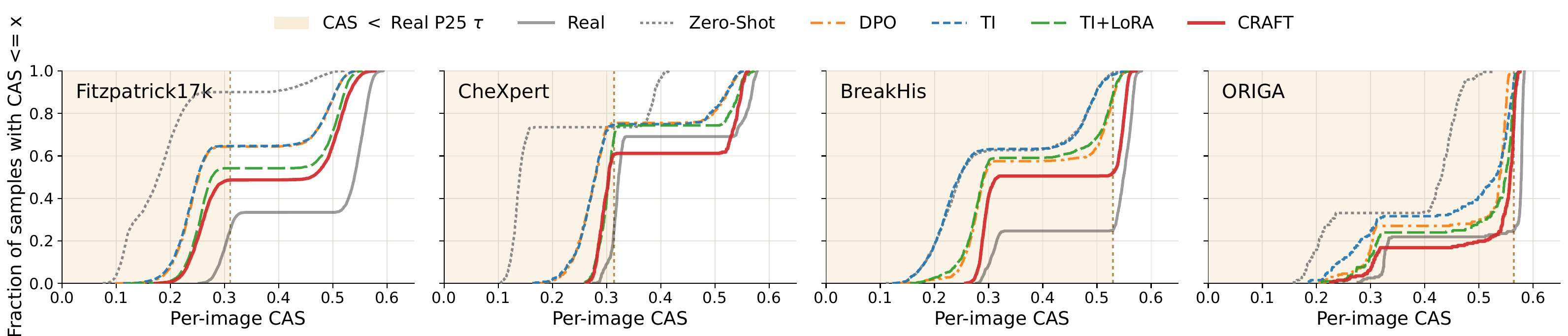}
  \caption{
  Empirical CDFs of per-image CAS.
  The shaded region denotes CAS below $\tau$, where $\tau$ is the 25th percentile of the real-image CAS distribution for each dataset. 
  Lower curves in the shaded region indicate fewer low-alignment generations. 
  CRAFT consistently reduces the lower-tail region across all four datasets.
  }
  \label{fig:cas_cdf}
  \vspace{-15pt}
\end{figure*}

Average CAS can hide clinically important failure cases, so we examine the lower tail directly. For each dataset, we set $\tau$ to the 25th percentile of the real-image CAS distribution; generated samples below $\tau$ fall below the lower quartile of real-image alignment under CAS.
Figure~\ref{fig:cas_cdf} shows that CRAFT shifts the per-image CAS distribution toward higher values and reduces the low-alignment region across all four datasets. 
CRAFT reduces the low-CAS rate relative to the strongest baseline from 54.2\% to 48.7\% on Fitzpatrick17k, 68.9\% to 60.8\% on CheXpert, 86.8\% to 52.1\% on BreakHis, and 89.8\% to 77.0\% on ORIGA, corresponding to an average relative reduction of 20.4\%.
CRAFT therefore improves not only mean alignment but also tail reliability.
Full per-method tail statistics are provided in Appendix~\ref{app:tail_analysis}.

\subsubsection{Effect of prompt enrichment}
\begin{wraptable}{r}{0.48\linewidth}
\vspace{-30pt}
\centering
\small
\setlength{\tabcolsep}{4pt}
\caption{Prompt-only enrichment ablation.}
\label{tab:prompt_only_baseline_main}
\begin{tabular}{lccc}
\toprule
Dataset & TI+LoRA & +$t_{ep}$ & \textbf{CRAFT} \\
\midrule
Fitzpatrick17k & .376 & .390 & \textbf{.417} \\
CheXpert       & .354 & .365 & \textbf{.401} \\
BreakHis       & .402 & .406 & \textbf{.417} \\
ORIGA          & .497 & .501 & \textbf{.515} \\
\bottomrule
\end{tabular}
\vspace{-10pt}
\end{wraptable}

To isolate prompt enrichment from reward optimization, Table~\ref{tab:prompt_only_baseline_main} compares TI+LoRA, TI+LoRA+$t_{ep}$, and CRAFT. 
The prompt-only variant improves over TI+LoRA on every dataset, confirming that MLLM-derived descriptions provide a stronger conditioning signal than sparse labels alone. However, CRAFT remains consistently higher than TI+LoRA+$t_{ep}$. This supports that reward optimization contributes beyond semantic prompt enrichment.
\begin{table}[t]
\centering
\footnotesize
\caption{
Additional validation evidence.
(a) CRAFT remains strongest under MetaCLIP2 and does not fall into the lowest-diversity regime.
(b) CRAFT achieves the highest structured checklist pass rate under an external medical evaluator.
}
\vspace{-4pt}

\begin{minipage}[t]{0.57\linewidth}
\centering
\textbf{(a) Evaluator robustness and diversity}\\[-1pt]
\setlength{\tabcolsep}{3.2pt}
\begin{tabular}{llccc}
\toprule
Dataset & Method & SigLIP CAS & MetaCLIP2 CAS & LPIPS \\
\midrule
\multirow{4}{*}{CheXpert}
& TI & 0.320 & 0.356 & 0.528 \\
& DPO & 0.320 & 0.355 & 0.529 \\
& TI+LoRA & 0.354 & 0.387 & 0.466 \\
& \textbf{CRAFT} & \textbf{0.401} & \textbf{0.417} & 0.510 \\
\midrule
\multirow{4}{*}{Fitzpatrick}
& TI & 0.327 & 0.368 & 0.559 \\
& DPO & 0.331 & 0.371 & 0.562 \\
& TI+LoRA & 0.376 & 0.413 & 0.543 \\
& \textbf{CRAFT} & \textbf{0.417} & \textbf{0.456} & 0.552 \\
\bottomrule
\end{tabular}
\end{minipage}
\hspace{0.025\linewidth}
\begin{minipage}[t]{0.38\linewidth}
\centering
\textbf{(b) Checklist audit}\\[-1pt]
\setlength{\tabcolsep}{4pt}
\begin{tabular}{lccc}
\toprule
Dataset & Zero-shot & TI+LoRA & \textbf{CRAFT} \\
\midrule
Fitzpatrick & 0.199 & 0.580 & \textbf{0.586} \\
CheXpert & 0.066 & 0.476 & \textbf{0.566} \\
BreakHis & 0.638 & 0.692 & \textbf{0.811} \\
ORIGA & 0.892 & 0.912 & \textbf{0.996} \\
\bottomrule
\end{tabular}
\end{minipage}

\label{tab:additional_validation}
\vspace{-5pt}
\end{table}

\subsubsection{Qualitative analysis}
Appendix~\ref{app:qualitative} provides qualitative comparisons across Fitzpatrick17k, CheXpert, BreakHis, and ORIGA. In Figure~\ref{fig:qualitative_main}, CRAFT more often expresses diagnosis-consistent visual patterns than the baselines, including melanoma border irregularity and color variegation, target-like erythema multiforme morphology, pneumonia-like localized opacity, cardiomegaly-related cardiac enlargement, histopathology tissue structure, and optic-disc morphology. These examples are qualitative support rather than definitive clinical evidence. Appendix~\ref{app:error_analysis} further reports representative CheXpert failure cases, including missing fine-grained pathology cues, diagnosis-inconsistent morphology, and clinically incomplete outputs.

\subsubsection{Downstream utility in a real and synthetic training setting}
We test whether improved alignment translates into practical utility. In a real and synthetic augmentation setting with 20\% synthetic samples per batch, following \citet{wang2025doctor}, CRAFT achieves the best downstream accuracy across all four datasets and three data scales (Figure~\ref{fig:downstream_aug}).
This includes full-data gains over the best baseline on Fitzpatrick17k (0.664 vs.\ 0.656), CheXpert (0.570 vs.\ 0.566), BreakHis (0.838 vs.\ 0.795), and ORIGA (0.801 vs.\ 0.791). We do not treat classifier improvement as standalone evidence of medical correctness, since synthetic images can aid training while still containing artifacts. Instead, downstream utility provides complementary evidence that the CAS gains are not merely evaluator-specific. Appendix~\ref{app:cas_correlation} further shows that method-level CAS is strongly correlated with downstream accuracy and macro-F1 on Fitzpatrick17k and CheXpert.


\subsubsection{Physician preference evaluation}
\begin{wrapfigure}{r}{0.49\linewidth}
  \vspace{-30pt}
  \centering
  \includegraphics[width=\linewidth]{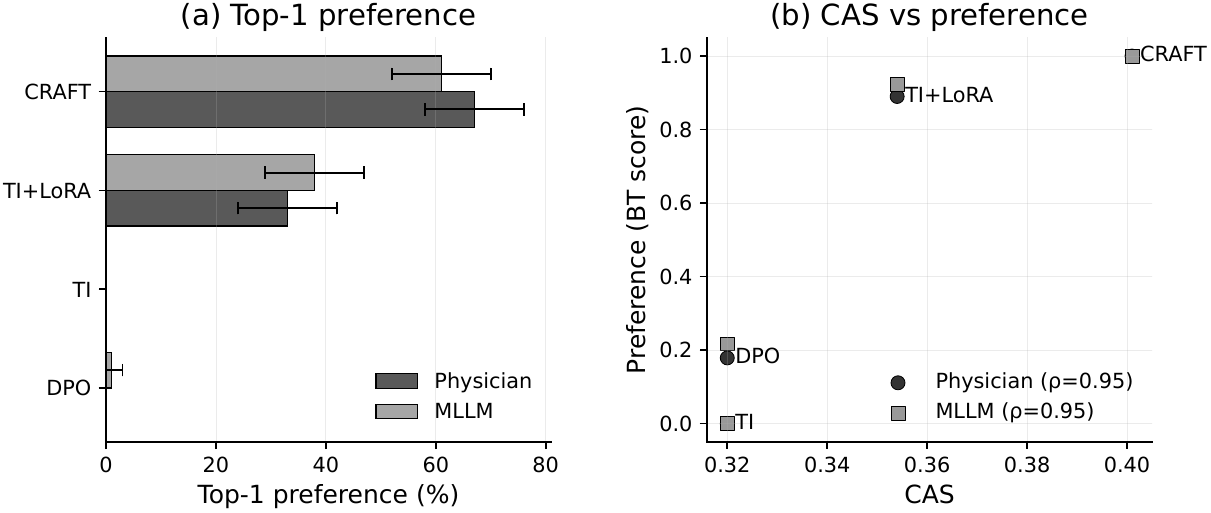}
  \caption{
  Physician preference evaluation on CheXpert. Two physicians ranked 100 randomized cases. CRAFT has the highest top-1 preference rate (67\%), and CAS correlates with Bradley--Terry preference scores.
  }
  \label{fig:chexpert_preference}
  \vspace{-15pt}
\end{wrapfigure}
To complement automated metrics, we conducted a blinded physician preference study on CheXpert. Two physicians independently ranked synthetic images from different methods across 100 randomly selected cases. As shown in Figure~\ref{fig:chexpert_preference}, CRAFT is ranked first in 67\% of cases, and CAS is strongly associated with Bradley--Terry preference scores derived from physician rankings. We frame this as supporting evidence rather than definitive clinical validation, given the limited scope (one modality, two raters).

\subsubsection{Additional validation: evaluator robustness, diversity, and checklist auditing}
\label{sec:eval_robustness}
Three concerns about CAS gains require evidence: evaluator overfitting, reduced sample diversity, and incomplete diagnostic content. We address each in turn.
Table~\ref{tab:additional_validation} provides two complementary checks. First, we evaluate with MetaCLIP2, an out-of-family image--text encoder not used during training, and report prompt-level LPIPS over 4 samples for each of 100 held-out prompts. Method ranking is unchanged under MetaCLIP2 on both datasets, and CRAFT retains the strongest alignment. LPIPS also shows no evidence of severe mode collapse: CRAFT does not fall into the lowest-diversity regime among trained methods. Second, structured checklist auditing with Lingshu \citep{xu2025lingshu}, an external medical evaluator not used in training, shows that CRAFT achieves the highest pass rate across all four datasets, with clear gains on CheXpert (+0.090 over TI+LoRA) and BreakHis (+0.119). We also compare against recent zero-shot general-purpose generators in Appendix~\ref{app:zeroshot_generators}. CRAFT outperforms the strongest zero-shot reference, Flux, on both Fitzpatrick17k (0.417 vs.\ 0.213 CAS) and CheXpert (0.401 vs.\ 0.305 CAS). Finally, because CRAFT uses a reference-anchored SFS reward during training, we test for training-set memorization in Appendix~\ref{app:memorization}. Across SSIM/pHash duplicate detection and DINOv2 \citep{oquab2023dinov2} nearest-neighbor analysis, we find no evidence of pixel-level or feature-space memorization.

\begin{table}[t]
\centering
\footnotesize
\renewcommand{\arraystretch}{0.88}
\caption{Ablation on Fitzpatrick17k. Left: adding a single reward to the base model. Right: leave-one-out from full CRAFT. Each component primarily affects its intended primitive.}
\vspace{-4pt}

\noindent
\begin{minipage}[t]{0.485\linewidth}
\centering
\textbf{(a) Single-component addition}\\
\resizebox{\linewidth}{!}{%
\begin{tabular}{lccccc}
\toprule
Variant & VDC & CCS & DD & SFS & CAS \\
\midrule
Base & .094 & .150 & .449 & .809 & .376 \\
+ VDC & \textbf{.164} & .145 & .468 & .813 & .398 \\
+ CCS & .128 & \textbf{.156} & .507 & .813 & .401 \\
+ DD  & .136 & .149 & \textbf{.514} & .812 & .403 \\
+ SFS & .130 & .148 & .470 & \textbf{.820} & .392 \\
\bottomrule
\end{tabular}
}
\end{minipage}%
\hfill%
\begin{minipage}[t]{0.485\linewidth}
\centering
\textbf{(b) Leave-one-out}\\
\resizebox{\linewidth}{!}{%
\begin{tabular}{lccccc}
\toprule
Variant & VDC & CCS & DD & SFS & CAS \\
\midrule
\textbf{Full} & \textbf{.166} & \textbf{.158} & \textbf{.520} & \textbf{.824} & \textbf{.417} \\
w/o DD & .161 & .156 & .471 & .815 & .401 \\
w/o Diff. & .163 & .153 & .492 & .819 & .407 \\
w/o VDC & .151 & .156 & .517 & .817 & .410 \\
w/o CCS & .164 & .146 & .510 & .823 & .411 \\
w/o SFS & .164 & .158 & .520 & .820 & .416 \\
\bottomrule
\end{tabular}%
}
\end{minipage}

\label{tab:ablation}
\vspace{-8pt}
\end{table}
\renewcommand{\arraystretch}{1.0}

\subsection{Ablation study}
\label{sec:ablation}
Each reward component primarily affects its intended primitive (Table~\ref{tab:ablation}).
In the incremental setting, adding VDC, CCS, DD, and SFS most strongly improves VDC, CCS, DD, and SFS, respectively. 
In the leave-one-out setting, removing DD causes the largest DD drop (0.520$\rightarrow$0.471), removing VDC causes the largest VDC drop (0.166$\rightarrow$0.151), and removing CCS causes the largest CCS drop (0.158$\rightarrow$0.146).
Aggregate CAS shifts are smaller because CAS averages four primitives. The pattern supports the reward decomposition: the components are complementary, not redundant. Appendix~\ref{app:dd_weight} further shows that performance is stable across a broad range of DD reward weights.

\paragraph{Additional appendix analyses.}
Several additional analyses support the main findings. 
Appendices~\ref{app:physician_pref}--\ref{app:auto_pref} provide physician and MLLM preference protocols, Appendix~\ref{app:fewshot} reports few-shot results, Appendices~\ref{app:hyperparam}--\ref{app:dd_weight} analyze hyperparameter and DD-weight sensitivity, and Appendix~\ref{app:label_dist} reports label distributions used for metric-based evaluation.

\section{Limitations}
Several limitations remain. First, both reward optimization and evaluation depend on pretrained foundation models; encoder biases or coverage gaps may affect scoring even with our separation between MedSigLIP (training reward) and SigLIP/MetaCLIP2 (evaluation).
Second, CAS is a proxy for clinical alignment, not a substitute for expert review. 
Third, SFS requires paired reference images during training and reference-based evaluation. Our memorization analysis finds no evidence of training-image copying but does not constitute a formal privacy guarantee, which would require differential-privacy or membership-inference analysis.
Finally, physician evaluation is limited to CheXpert and two medical doctors.

\section{Conclusions}
We introduced CRAFT, a reward-aligned finetuning framework that adapts diffusion models toward clinically aligned medical image synthesis. 
CRAFT combines label-driven semantic enrichment with differentiable rewards for visual description consistency, clinical criteria satisfaction, diagnostic discriminability, and semantic feature similarity.
Across four medical imaging domains, CRAFT improves average CAS, downstream classification performance, and the low-alignment tail of per-image CAS, indicating fewer hallucination-like or clinically implausible generations under our proxy metric.
Together with out-of-family evaluator analysis, diversity measurement, structured checklist auditing, and physician preference results on CheXpert, these findings indicate that optimizing clinically motivated reward components is a productive path toward reliable medical image generation.




\bibliographystyle{plainnat}
\bibliography{main}

\newpage
\appendix
\section*{}
\begin{center}
    {\Large \textbf{Appendix}}
\end{center}
\vspace{0.2in}
\hrule
{\large \noindent\textbf{Table of Contents}}

\begin{itemize}
    \item \textbf{A. Implementation details and reproducibility}
    \begin{itemize}
        \item A.1 Clinical alignment module (CAM)
        \item A.2 Diagnostic discriminability linear probe
        \item A.3 Statistical uncertainty
        \item A.4 Compute resources
        \item A.5 Code and data access
    \end{itemize}

    \item \textbf{B. Prompt enrichment pipeline}
    \begin{itemize}
        \item B.1 Label conditioning to prevent diagnosis hallucination
        \item B.2 Schema enforcement across domains
    \end{itemize}

    \item \textbf{C. Clinical checklists}
    \begin{itemize}
        \item C.1 Fitzpatrick17k clinical checklists
        \item C.2 CheXpert clinical checklists
        \item C.3 BreakHis clinical checklists
        \item C.4 ORIGA clinical checklists
    \end{itemize}
    
    \item \textbf{D. Physician preference evaluation}
    \item \textbf{E. Automated preference-based evaluation}
    \item \textbf{F. Few-shot adaptation}
    \item \textbf{G. CAS--downstream utility correlation}
    \item \textbf{H. Low-alignment tail analysis}
    \item \textbf{I. Hyperparameter sensitivity analysis}
    \item \textbf{J. Diagnostic discriminability weight sensitivity}
    \item \textbf{K. Extended qualitative results}
    \item \textbf{L. Comparison with general-purpose zero-shot generators}
    \item \textbf{M. Memorization analysis}
    \item \textbf{N. Dataset label distributions}
    \item \textbf{O. Error analysis on CheXpert}
    \item \textbf{P. Existing assets and licenses}
    \item \textbf{Q. Broader impact}
\end{itemize}

\hrule

\vspace{0.2in}

\section{Implementation details and reproducibility}
\label{app:reproducibility}
\subsection{Clinical alignment module (CAM)}
The CAM uses a frozen medical VLM as a differentiable critic. We employ MedSigLIP-448 (\url{https://huggingface.co/google/medsiglip-448}) as the backbone VLM. All image and text embeddings used in Eq.~(2--5) are extracted from this model. The VLM is kept fully frozen throughout CRAFT training and is used only to compute reward signals. No gradients are backpropagated into the VLM parameters. Given a generated image and its associated textual inputs, CAM outputs four scalar rewards corresponding to visual description consistency, clinical criteria satisfaction, diagnostic discriminability, and semantic feature similarity.

\subsection{Diagnostic discriminability linear probe}
The diagnostic integrity reward $r_\textrm{dd}$ (Eq.~4) is implemented using a linear probe trained on frozen VLM embeddings. A single linear layer is trained using only the real training split of each dataset. We use Adam optimizer with learning rate $1\times10^{-3}$ for 50 epochs and batch size of 256. During CRAFT optimization, the linear layer is kept frozen and used only to compute reward gradients with respect to the diffusion model. During training, the probe provides log-likelihood rewards for stable gradient optimization, while we report classification accuracy as a diagnostic discriminability metric during evaluation. We train two diagnostic linear probes with the same protocol but on different embedding spaces: a MedSigLIP-space probe for the differentiable DD reward during CRAFT training, and a SigLIP-space probe for metric computation. Both probes are trained only on real training images and kept frozen during generation or evaluation.

\subsection{Statistical uncertainty}
\label{app:statistical_uncertainty}
For synthetic-method CAS, we report mean and standard deviation over five random seeds. The main table reports means for compactness, and Table~\ref{tab:cas_uncertainty} provides the corresponding CAS uncertainty where available. For physician preference evaluation, we report 95\% bootstrap confidence intervals. These uncertainty estimates quantify run-to-run variability and evaluator-sampling variability, not clinical uncertainty.

\begin{table}[t]
\centering
\small
\caption{CAS uncertainty over five seeds. Values are mean $\pm$ standard deviation. ORIGA standard deviations are reported when seed-level logs are available.}
\label{tab:cas_uncertainty}
\begin{tabular}{lcccc}
\toprule
Method & Fitzpatrick17k & CheXpert & BreakHis & ORIGA \\
\midrule
Zero-Shot & $.202{\pm}.004$ & $.205{\pm}.006$ & $.294{\pm}.003$ & $.363{\pm}.007$ \\
DPO       & $.331{\pm}.004$ & $.320{\pm}.004$ & $.378{\pm}.005$ & $.462{\pm}.005$ \\
TI        & $.327{\pm}.003$ & $.320{\pm}.004$ & $.367{\pm}.003$ & $.456{\pm}.004$ \\
TI+LoRA   & $.376{\pm}.003$ & $.354{\pm}.005$ & $.402{\pm}.002$ & $.497{\pm}.005$ \\
CRAFT     & $.417{\pm}.002$ & $.401{\pm}.003$ & $.417{\pm}.003$ & $.515{\pm}.004$ \\
\bottomrule
\end{tabular}
\end{table}

\subsection{Compute resources}
\label{app:compute}
All experiments were run on NVIDIA RTX 6000 Ada GPUs. Table~\ref{tab:compute} summarizes the approximate compute used for the main experiment families. Runtimes vary by dataset size, number of generated samples, and evaluator choice; when exact wall-clock logs were unavailable, we estimated runtimes from checkpoint or output-file timestamps.

\begin{table*}[t]
\centering
\small
\setlength{\tabcolsep}{4pt}
\renewcommand{\arraystretch}{1.08}
\caption{Approximate compute used for the main experiment families. GPU-hours are approximate and are reported to indicate scale rather than exact accounting.}
\label{tab:compute}
\resizebox{\linewidth}{!}{
\begin{tabular}{p{3.0cm} p{4.4cm} p{2.3cm} p{4.4cm} p{2.0cm} p{2.0cm}}
\toprule
Experiment family & Scope & Hardware & Approximate runtime & Seeds & Approx. GPU-hours \\
\midrule

Textual inversion
& Disease-token learning for class-conditioned generation
& 1 GPU per class
& $\sim$0.9 h per class on CheXpert
& 1
& $\sim$3.6 \\

TI+LoRA finetuning
& Standard adaptation baseline and prompt-enriched variant
& 1 GPU per run
& $\sim$10--11 min per CheXpert run
& 1
& $\sim$0.4 \\

CRAFT finetuning
& Reward-aligned finetuning and component / hyperparameter sweeps
& 1 GPU per run
& $\sim$1.3 h per CheXpert run; $\sim$8--9 h per Fitzpatrick17k run
& train seed 42; evaluation over 5 seeds
& $\sim$50--55 \\

CAS / evaluator scoring
& SigLIP CAS scoring, diagnostic-probe evaluation, and MetaCLIP2 reevaluation
& 1 GPU
& $\sim$1 min per CheXpert seed; $\sim$5 min per Fitzpatrick17k seed
& 5
& $<1$ for main tables \\

Downstream augmentation
& Real+synthetic classifier training across datasets and data scales
& 1 GPU per run
& $\sim$1--2 min per classifier run
& 1--3 depending on setting
& $\sim$1--2 \\

Checklist auditing
& External medical-evaluator checklist pass-rate analysis
& 1 GPU per run
& $\sim$2.5--3 h per audit run
& N/A
& $\sim$8--9 \\

Preference evaluation
& MLLM-based candidate ranking and physician-preference aggregation
& 1 GPU for MLLM judging
& $\sim$7--15 min for MLLM judging batches
& N/A
& $<1$ for MLLM judging \\

\bottomrule
\end{tabular}
}
\end{table*}

\subsection{Code and data access}
\label{app:code_data}
We release anonymized code at \url{https://anonymous.4open.science/r/CRAFT-07B4}. The repository includes training scripts, evaluation scripts, configuration files, and table/figure generation code. Raw medical images are not redistributed. Users should obtain Fitzpatrick17k, CheXpert, BreakHis, and ORIGA from their original sources and comply with the corresponding licenses or data-use agreements. We provide dataset preprocessing instructions, split definitions where redistribution is permitted, and scripts for regenerating enriched prompts, checklists, synthetic images, and CAS metrics.

\section{Prompt enrichment pipeline} \label{app:prompt}
To ensure the generated enriched prompts ($t_{ep}$) and checklists ($t_c$) are consistent and high-quality, we implement a strict generation pipeline using GPT-4o \citep{achiam2023gpt}. The process is governed by two key mechanisms: Label Conditioning and Schema Enforcement.

\subsubsection{Label conditioning to prevent hallucination}
A common failure mode in medical VLM captioning is ungrounded diagnosis inference, where the model hallucinates a condition based on visual effects. To mitigate this, we provide the ground-truth label $y$ as a hard constraint in the system prompt. The model is instructed to describe the visual sign of the provided label $y$ within the image $x$, rather than attempting to predict $y$. This encourages the generated descriptions to remain consistent with the provided label and reduces unsupported diagnostic inference, although the resulting descriptions remain automatically generated and may still contain imperfect visual details.

\subsubsection{Schema enforcement}
We enforce strict output schemas to standardize the vocabulary across the dataset and filter out malformed responses. 
\begin{itemize}
    \item \textbf{Dermatology (Fitzpatrick17k):} We constrain the \textit{body part} and \textit{skin type} fields to a closed set of values (e.g., fair, brown, dark) to prevent uncontrolled vocabulary expansion and long-tail descriptor fragmentation.
    \item \textbf{Histopathology (BreakHis):} We constrain tissue context and key findings to a fixed schema tailored to breast histopathology morphology at 100$\times$ magnification. This reduces unsupported free-form pathology language and keeps descriptions focused on patch-visible architectural and cytologic features.
    \item \textbf{Radiology (CheXpert):} We use label-dependent logic to prevent pathology leakage. For example, if the ground truth is "No Finding", the system prompt explicitly forbids the generation of terms like "opacity" or "consolidation", ensuring the caption remains "clear lungs with sharp costophrenic angles."
    \item \textbf{Ophthalmology (ORIGA):} We constrain optic-disc findings to a fixed glaucoma-oriented schema, including cup size, neuroretinal rim status, vessel course, and peripapillary changes. This prevents free-form ophthalmology language and keeps descriptions focused on visible fundus features relevant to glaucomatous optic neuropathy.
\end{itemize}

\begin{tcolorbox}[
    colback=gray!10,       
    colframe=gray!60,      
    title=\textbf{System Prompt Logic for Dermatology (Fitzpatrick17k)}, 
    sharp corners=south,   
    boxrule=0.5mm
]
\ttfamily \small 
You're an expert dermatologist. You extract visual morphology from ONE clinical image.
The ground truth diagnosis is provided below as \textbf{\{label\}}. Do not invent findings outside of this diagnosis.

\vspace{0.5em}
\textbf{Return STRICTLY valid JSON with keys:}
\begin{itemize}
    \item body\_part (MUST be one of: \{allowed\_body\_parts\})
    \item lesion\_features (concise phrase $\le$ 20 words: shape, texture, color)
\end{itemize}

\textbf{Style constraints:}
\begin{itemize}
    \item Use plain language. Avoid medical jargon (say "red" not "erythematous").
    \item Focus on visual attributes of \textbf{\{label\}} (shape, border, color).
    \item Output JSON only. No markdown.
\end{itemize}
\end{tcolorbox}

\begin{tcolorbox}[
    colback=gray!10,
    colframe=gray!60,
    title=\textbf{System Prompt Logic for Histopathology (BreakHis)},
    sharp corners=south,
    boxrule=0.5mm
]
\ttfamily \small
You're an expert breast pathologist. You describe ONE H\&E breast histopathology patch at 100x magnification.

\vspace{0.5em}
The ground-truth diagnosis is provided below as \textbf{\{label\}}.
Do not predict a diagnosis.
Do not invent findings outside of this diagnosis.
Only mention morphology that is reasonably visible in a single 100x patch.

\vspace{0.5em}
\textbf{Return STRICTLY valid JSON with keys:}
\begin{itemize}
    \item tissue\_context
    \item key\_findings
    \item visual\_summary
\end{itemize}

\textbf{Schema rules:}
\begin{itemize}
    \item tissue\_context: MUST be one of \\
    \texttt{["glandular", "fibroepithelial", "stromal", "papillary", "mucinous", "infiltrative", "mixed"]}
    \item key\_findings: comma-separated list chosen ONLY from \textbf{\{allowed\_key\_findings\}}
    \item visual\_summary: concise phrase $\leq$ 18 words
\end{itemize}

\textbf{Style constraints:}
\begin{itemize}
    \item Use visual morphology only.
    \item Prefer short pathology terms such as \\
    "enlarged acini", "slit-like ducts", "leaf-like stromal fronds", \\
    "tightly packed tubules", ``irregular ducts", "single-file cells", \\
    "mucin pools", "papillary fronds", and "fibrovascular cores".
    \item Do not mention biomarkers, grade, mitotic rate, receptor status, prognosis, or treatment.
    \item Do not mention features that cannot be supported by the patch.
    \item Avoid long sentences and differential diagnosis language.
    \item Output JSON only. No markdown.
\end{itemize}
\end{tcolorbox}

\begin{tcolorbox}[
    colback=gray!10,       
    colframe=gray!60,      
    title=\textbf{System Prompt Logic for Radiology (CheXpert)}, 
    sharp corners=south,   
    boxrule=0.5mm
]
\ttfamily \small 
You're an expert radiologist. Describe a frontal chest x-ray using simple visual language.
The ground truth diagnosis is provided below as \textbf{\{label\}}. Do not invent findings outside of this diagnosis.

\textbf{Return STRICTLY valid JSON with keys:}
\begin{itemize}
    \item devices (e.g., "pacemaker", "picc line", or "none")
    \item key\_findings (comma-separated list from allowed pool)
    \item visual\_summary (concise phrase $\le$ 16 words)
\end{itemize}

\textbf{Style constraints:}
\begin{itemize}
    \item Avoid Negations: Do NOT say "no pleural effusion". Instead, say "sharp costophrenic angles".
    \item Prefer positive descriptors (e.g., "clear lungs") for normal regions.
    \item If label="No Finding", forbid pathology terms entirely.
    \item Output JSON only. No markdown.
\end{itemize}
\end{tcolorbox}

\begin{tcolorbox}[
    colback=gray!10,
    colframe=gray!60,
    title=\textbf{System Prompt Logic for Ophthalmology (ORIGA)},
    sharp corners=south,
    boxrule=0.5mm
]
\ttfamily \small
You're an expert ophthalmologist. You describe ONE retinal fundus image.

\vspace{0.5em}
The ground-truth diagnosis is provided below as \textbf{\{label\}}.
Do not predict a diagnosis.
Do not invent findings outside of this diagnosis.
Only mention morphology that is reasonably visible in a single fundus photograph.

\vspace{0.5em}
\textbf{Return STRICTLY valid JSON with keys:}
\begin{itemize}
    \item disc\_appearance
    \item key\_findings
    \item visual\_summary
\end{itemize}

\textbf{Schema rules:}
\begin{itemize}
    \item disc\_appearance: MUST be one of \\
    \texttt{["normal disc", "enlarged cup", "rim thinning", "notched rim", "glaucomatous disc"]}
    \item key\_findings: comma-separated list chosen ONLY from \textbf{\{allowed\_key\_findings\}}
    \item visual\_summary: concise phrase $\leq$ 18 words
\end{itemize}

\textbf{Style constraints:}
\begin{itemize}
    \item Use only visible fundus morphology.
    \item Focus on optic cup size, neuroretinal rim, vessel course, and peripapillary changes.
    \item Do not mention visual field loss, intraocular pressure, OCT, prognosis, or treatment.
    \item Do not mention features not clearly supported by the image.
    \item Output JSON only. No markdown.
\end{itemize}
\end{tcolorbox}


\section{Clinical checklists} \label{app:checklists}
\subsection{Fitzpatrick17k clinical checklists} \label{app:fitz_checklist}
Table \ref{tab:fitz_checklist} details the structured visual criteria used to guide the Clinical Alignment Module (CAM). We decompose complex dermatological pathology into five fundamental visual attributes (Location, Lesion Type, Shape/Size, Color, Texture) to ensure interpretability by the general-domain VLMs. These attributes were selected to balance clinical completeness with interpretability by general-domain vision–language models.

\begin{longtable}{p{0.2\textwidth} p{0.2\textwidth} p{0.5\textwidth}}
    \caption{Structured Clinical Criteria for Fitzpatrick17k Classes.} \label{tab:fitz_checklist} \\
    \toprule
    \textbf{Condition} & \textbf{Attribute} & \textbf{Visual Description} \\
    \midrule
    \endfirsthead
    

    \midrule
    \multicolumn{3}{r}{Continued on next page} \\
    \bottomrule
    \endfoot

    \bottomrule
    \endlastfoot

    \textbf{Acne}
    & Location & Face, forehead, chest, shoulders, upper back \\
    & Lesion Type & Bumps including comedones (whiteheads, blackheads) and inflamed pimples \\
    & Shape/Size & Small clogged-pore bumps; larger tender nodules/cysts \\
    & Color & Red or skin-colored bumps; blackheads have dark plug \\
    & Texture & Oily or shiny skin; pus or crust if ruptured \\
    \midrule

    \textbf{Actinic Keratosis}
    & Location & Sun-exposed areas (face, scalp, ears, hands) \\
    & Lesion Type & Rough, scaly patch or small crusty bump \\
    & Shape/Size & Flat or slightly raised, under 2.5 cm \\
    & Color & Pink, red, or brownish; yellowish crust \\
    & Texture & Dry, coarse, sandpaper-like surface \\
    \midrule

    \textbf{Allergic Contact Dermatitis}
    & Location & Contact sites (hands, face, neck) \\
    & Lesion Type & Red patches with blisters or swelling \\
    & Shape/Size & Irregular shape following exposure \\
    & Color & Pink to red; purple/brown on dark skin \\
    & Texture & Weepy, crusty, or scaly \\
    \midrule

    \textbf{Basal Cell Carcinoma} 
    & Location & Sun-exposed areas (face, nose, ears, neck, scalp, shoulders) \\
    & Lesion Type & Pearly or waxy bump/nodule, or flat scaly patch with a raised edge \\
    & Shape/Size & Small, round/oval; can ulcerate or develop a central depression \\
    & Color & Translucent or pearly on fair skin; brown/black or glossy dark on darker skin \\
    & Texture & Smooth, shiny surface; can crust or scab with central ulceration \\
    \midrule

    \textbf{Eczema} 
    & Location & Flexural areas (inner elbows, behind knees), hands, neck, eyelids \\
    & Lesion Type & Patches or plaques, sometimes with small blisters or bumps \\
    & Shape/Size & Ill-defined patches varying in size; often bilateral or symmetric \\
    & Color & Red or pink on lighter skin; purple, gray, or dark brown on darker skin \\
    & Texture & Dry, flaky, or scaly; can become thick and leathery (lichenification) \\
    \midrule

    \textbf{Erythema Multiforme}
    & Location & Hands, feet, arms, legs; mucous membranes (lips, mouth) \\
    & Lesion Type & Target (bull's-eye) lesions with concentric rings \\
    & Shape/Size & Round lesions (1--3 cm) with dark center, pale ring, and outer red ring \\
    & Color & Dark red/purple center, pale ring, red outer zone; gray/dark on dark skin \\
    & Texture & Mostly flat but can have a blistered or raised center \\
    \midrule

    \textbf{Folliculitis}
    & Location & Hair-bearing areas (beard, scalp, underarms, legs, buttocks) \\
    & Lesion Type & Small pustules or red papules centered around hair follicles \\
    & Shape/Size & Clusters of 2--5 mm bumps; each usually pierced by a hair \\
    & Color & Red/pink (light skin); dark/hyperpigmented (dark skin); white/yellow pus \\
    & Texture & Dome-shaped, often fluid-filled top; crusts if ruptured \\
    \midrule
    
    \textbf{Granuloma Annulare}
    & Location & Hands, feet, wrists, ankles; occasionally trunk/limbs \\
    & Lesion Type & Smooth, firm bumps (papules) forming rings \\
    & Shape/Size & Annular (ring-shaped) up to few cm; composed of small papules \\
    & Color & Skin-colored, pink, or reddish; purple/brown on darker skin \\
    & Texture & Generally smooth surface; little to no flaking or scale \\
    \midrule
    
    \textbf{Keloid}
    & Location & Chest, shoulders, earlobes, jawline, or sites of injury \\
    & Lesion Type & Overgrown scar tissue extending beyond original wound \\
    & Shape/Size & Raised, irregular shape; can grow large and claw-like \\
    & Color & Pink/red (light skin); dark brown, purple, or black (dark skin) \\
    & Texture & Smooth, hairless, firm, rubbery, or shiny surface \\
    \midrule
    
    \textbf{Lichen Planus}
    & Location & Wrists, forearms, ankles, lower back, mucous membranes \\
    & Lesion Type & Flat-topped papules; can form plaques \\
    & Shape/Size & Polygonal, 2--10 mm papules \\
    & Color & Violaceous (purple); gray-brown or hyperpigmented on dark skin \\
    & Texture & Shiny surface with fine white lines (Wickham's striae) \\
    \midrule

    \textbf{Lupus Erythematosus}
    & Location & Face (butterfly rash), scalp, ears, sun-exposed areas \\
    & Lesion Type & Malar rash (flat/raised); discoid lesions (scaly/scarred) \\
    & Shape/Size & Butterfly shape across nose/cheeks; discoid are coin-shaped \\
    & Color & Pink-red; can be dark red or hyperpigmented on darker skin \\
    & Texture & Smooth (malar) or rough/scaly with central scarring (discoid) \\
    \midrule

    \textbf{Melanoma} 
    & Location & Trunk, limbs, face, nails; in darker skin, often on palms/soles \\
    & Lesion Type & Atypical mole or patch; irregular shape and color \\
    & Shape/Size & Asymmetric, often $>$6 mm, with irregular or notched borders \\
    & Color & Multiple shades (brown, black, red, white, blue); variegated \\
    & Texture & Smooth early; may become raised, crusted, or ulcerated if advanced \\
    \midrule

    \textbf{Mycosis Fungoides}
    & Location & Non-sun-exposed areas (buttocks, thighs); can be widespread \\
    & Lesion Type & Patches (eczema-like), plaques (thick), or tumor nodules \\
    & Shape/Size & Irregular patches/plaques; tumors can be large \\
    & Color & Pink-red to reddish-brown; dark/hyperpigmented on dark skin \\
    & Texture & Dry, scaly, wrinkled (cigarette paper) patches; tumors smooth/ulcerated \\
    \midrule

    \textbf{Pityriasis Rosea}
    & Location & Trunk (back, chest, abdomen); "Christmas tree" pattern \\
    & Lesion Type & Single "Herald patch" followed by smaller oval lesions \\
    & Shape/Size & Herald patch (2--6 cm); daughter lesions (1--2 cm ovals) \\
    & Color & Pink/salmon (light skin); gray, brown, or violet (dark skin) \\
    & Texture & Fine "collarette" scale at the inner edge of lesions \\
    \midrule
    
    \textbf{Prurigo Nodularis}
    & Location & Arms, legs, upper back (areas reachable for scratching) \\
    & Lesion Type & Firm, extremely itchy nodules \\
    & Shape/Size & Dome-shaped nodules 1--3 cm; often multiple \\
    & Color & Pink, red, or skin-toned; often dark brown/black borders \\
    & Texture & Hard, warty, or crusted surface; often excoriated (scratched) \\
    \midrule

    \textbf{Psoriasis} 
    & Location & Elbows, knees, scalp, lower back; can affect nails, palms, soles \\
    & Lesion Type & Well-demarcated plaques with thick, scaly surface \\
    & Shape/Size & Round/oval or irregular plaques; range from small patches to large areas \\
    & Color & Pink/red with silvery scales (light skin); purple/brown with gray scales (dark skin) \\
    & Texture & Dry, flaky scales that can be peeled off; underlying skin may bleed \\
    \midrule

    \textbf{Sarcoidosis}
    & Location & Face, shins, scars \\
    & Lesion Type & Firm plaques or nodules \\
    & Shape/Size & Broad plaques; 1-5 cm nodules \\
    & Color & Red-brown; purple; discoloration \\
    & Texture & Smooth, firm, rubbery \\
    \midrule

    \textbf{Scabies}
    & Location & Finger webs, wrists, waist \\
    & Lesion Type & Linear burrows and small allergic papules/vesicles \\
    & Shape/Size & Burrows (5--15 mm wavy lines); small bumps (1--2 mm) \\
    & Color & Skin-toned, pink, or red; dark spots on darker skin \\
    & Texture & Burrows feel like slight ridges; crusting from scratching \\
    \midrule
    
    \textbf{Squamous Cell Carcinoma}
    & Location & Sun-exposed areas (face, ears, lips, hands), scars \\
    & Lesion Type & Crusty bump, ulcer, or thick plaque \\
    & Shape/Size & Firm nodule or patch; often $>$1 cm; can be crater-like \\
    & Color & Pink/red base; brown/dark on dark skin; white keratin \\
    & Texture & Rough, scaly, hyperkeratotic; may bleed or ulcerate \\
    \midrule

    \textbf{Vitiligo} 
    & Location & Face (eyes, mouth), hands, feet, arms, legs, genitals \\
    & Lesion Type & Depigmented patches with well-defined borders \\
    & Shape/Size & Irregular shapes; can start small and enlarge; often symmetrical \\
    & Color & Completely white or pale; high contrast against surrounding skin \\
    & Texture & Normal skin texture (no scaling or thickening), only color is lost \\
    \midrule
\end{longtable}

\subsection{CheXpert clinical checklists}
\label{app:chexpert_checklist}

For the radiological domain, we adapt the Color attribute to represent radiographic opacity and density. As shown in Table \ref{tab:chexpert_checklist}, we translate specific pathologies (e.g., Pleural Effusion) into geometric primitives (e.g., Meniscus sign, Homogeneous opacity) to maximize alignment with the VLM's visual priors.

\begin{table}[h]
\centering
\caption{Structured Clinical Criteria for CheXpert Classes.}
\label{tab:chexpert_checklist}
\begin{tabularx}{\textwidth}{l l X}
\toprule
\textbf{Condition} & \textbf{Attribute} & \textbf{Visual Description} \\
\midrule

\textbf{No Finding} 
& Location & Entire chest field (Lungs, Heart, Pleura) \\
& Lesion Type & Normal anatomy; no pathological opacities \\
& Shape/Size & Heart size normal ($<$50\% chest width); sharp triangular costophrenic angles \\
& Color & Lungs are dark (radiolucent); bones/heart are bright white \\
& Texture & Clear lung fields with fine vascular markings; no haziness \\
\midrule

\textbf{Cardiomegaly} 
& Location & Central chest (Mediastinum/Heart) \\
& Lesion Type & Enlargement of the cardiac silhouette \\
& Shape/Size & Globular or widened heart shadow; width $>$50\% of rib cage \\
& Color & Enlarged central white opacity \\
& Texture & Smooth, distinct heart borders; vascular congestion may be present \\
\midrule

\textbf{Pneumonia} 
& Location & Focal area in lungs (often asymmetric, lobar, or segmental) \\
& Lesion Type & Consolidation or Infiltrate (filled airspaces) \\
& Shape/Size & Patchy, irregular cloud-like shape; fluffy or ill-defined borders \\
& Color & Increased whiteness (opacity) against dark lung background \\
& Texture & Fluffy or hazy ``ground glass'' appearance; may show air bronchograms \\
\midrule

\textbf{Pleural Effusion} 
& Location & Lung bases (bottom corners) or lining the chest wall \\
& Lesion Type & Fluid accumulation masking the diaphragm \\
& Shape/Size & Meniscus sign (U-shaped curve); blunting of sharp costophrenic angle \\
& Color & Dense, homogeneous white opacity \\
& Texture & Smooth, uniform density (water-like); obscures the lung base \\
\midrule
\bottomrule
\end{tabularx}
\end{table}

\subsection{BreakHis clinical checklists}
\label{app:BreakHis_checklist}
For histopathology, we define structured visual criteria that describe tissue context, lesion type, architectural pattern, color, and texture. As shown in Table~\ref{tab:BreakHis_checklist}, these criteria are designed to capture class-defining morphology at 100$\times$ magnification while remaining interpretable to a vision--language critic. Compared with dermatology and radiology, the emphasis here is less on gross anatomy and more on glandular structure, stromal organization, and characteristic epithelial arrangements.

\begin{longtable}{p{0.22\textwidth} p{0.20\textwidth} p{0.50\textwidth}}
\caption{Structured Histopathology Criteria for BreakHis Classes.}
\label{tab:BreakHis_checklist} \\
\toprule
\textbf{Condition} & \textbf{Attribute} & \textbf{Visual Description} \\
\midrule
\endfirsthead

\midrule
\multicolumn{3}{r}{Continued on next page} \\
\bottomrule
\endfoot

\bottomrule
\endlastfoot

\textbf{Adenosis}
& Tissue Context & Lobulocentric glandular units with crowded acini in fibrous stroma \\
& Lesion Type & Benign proliferative glandular lesion with enlarged acini \\
& Shape/Size & Numerous small round acini with relatively regular lumina; clustered lobular pattern \\
& Color & Pink eosinophilic stroma with blue-purple uniform epithelial nuclei \\
& Texture & Orderly gland architecture; smooth outlines; bland cytology \\
\midrule

\textbf{Fibroadenoma}
& Tissue Context & Fibroepithelial lesion with ducts embedded in fibrous stroma \\
& Lesion Type & Well-circumscribed biphasic benign lesion \\
& Shape/Size & Round to slit-like ducts compressed by stroma; pushing contours \\
& Color & Pale pink collagenous stroma with evenly basophilic epithelial nuclei \\
& Texture & Smooth fibrous background; bland epithelium; low atypia \\
\midrule

\textbf{Phyllodes Tumor}
& Tissue Context & Fibroepithelial lesion with epithelial-lined stromal clefts \\
& Lesion Type & Leaf-like stromal fronds projecting into clefted spaces \\
& Shape/Size & Elongated branching fronds and cleft-like spaces; broader architecture than fibroadenoma \\
& Color & Pink stromal fronds with blue-purple epithelial and stromal nuclei \\
& Texture & Undulating leaf-like contours; more cellular stroma; mixed epithelial-stromal pattern \\
\midrule

\textbf{Tubular Adenoma}
& Tissue Context & Gland-forming lesion dominated by tightly packed tubules \\
& Lesion Type & Benign tubular epithelial proliferation \\
& Shape/Size & Numerous small round or oval tubules with open lumina; minimal intervening stroma \\
& Color & Pink scant stroma with blue-purple uniform nuclei around luminal spaces \\
& Texture & Dense orderly tubular pattern; smooth gland borders; bland cytology \\
\midrule

\textbf{Ductal Carcinoma}
& Tissue Context & Infiltrative epithelial lesion within fibrotic or desmoplastic stroma \\
& Lesion Type & Malignant duct-forming carcinoma \\
& Shape/Size & Irregular angulated ducts, nests, or cords with variable lumina \\
& Color & Hyperchromatic blue-purple nuclei in pink desmoplastic stroma \\
& Texture & Crowded pleomorphic cells; jagged infiltrative architecture; reduced uniformity \\
\midrule

\textbf{Lobular Carcinoma}
& Tissue Context & Discohesive infiltrating cells in fibrous stroma, often around ducts or lobules \\
& Lesion Type & Malignant lobular-type infiltrate \\
& Shape/Size & Single-file cords and targetoid periductal arrangement; little tubule formation \\
& Color & Small dark nuclei scattered through pale to pink stroma \\
& Texture & Loose discohesive pattern; subtle infiltrative spread; monotonous small cells \\
\midrule

\textbf{Mucinous Carcinoma}
& Tissue Context & Tumor cell clusters suspended within extracellular mucin pools \\
& Lesion Type & Mucin-producing carcinoma \\
& Shape/Size & Rounded clusters or strips of cells floating in large mucin lakes \\
& Color & Pale blue to lightly basophilic mucin with darker tumor cell nuclei \\
& Texture & Smooth gelatinous background; low-density floating cellular islands; soft lobulated spaces \\
\midrule

\textbf{Papillary Carcinoma}
& Tissue Context & Papillary epithelial proliferation organized around fibrovascular cores \\
& Lesion Type & Malignant papillary lesion \\
& Shape/Size & Branching papillary fronds with central fibrovascular stalks \\
& Color & Pink fibrovascular cores lined by blue-purple crowded epithelial cells \\
& Texture & Frond-like branching surface; layered epithelium; delicate core structures \\
\midrule
\end{longtable}

\begin{longtable}{p{0.22\textwidth} p{0.20\textwidth} p{0.48\textwidth}}
\caption{ORIGA Clinical Checklists.}
\label{tab:origa_checklist} \\
\toprule
\textbf{Condition} & \textbf{Attribute} & \textbf{Visual Description} \\
\midrule
\endfirsthead

\midrule
\multicolumn{3}{r}{Continued on next page} \\
\bottomrule
\endfoot

\bottomrule
\endlastfoot

\textbf{Normal}
    & Optic Disc & Small to moderate optic cup with balanced cup-to-disc ratio and preserved neuroretinal rim \\
    & Rim & Neuroretinal rim appears circumferentially intact without focal thinning or notching \\
    & Vessels & Retinal vessels near the disc follow a normal course without marked nasalization or bayoneting \\
    & Peripapillary Region & No obvious nerve fiber layer defect, disc hemorrhage, or glaucomatous peripapillary change \\
    & Overall Fundus & Overall retinal fundus appearance is consistent with a non-glaucomatous eye \\
    \midrule

    \textbf{Glaucoma}
    & Optic Disc & Enlarged optic cup or increased vertical cup-to-disc ratio is visible \\
    & Rim & Neuroretinal rim shows focal or diffuse thinning, notching, or clear rim loss, especially in the superior or inferior rim \\
    & Vessels & Retinal vessels near the disc show nasalization, bayoneting, or displacement around the cup edge \\
    & Peripapillary Region & Peripapillary nerve fiber layer loss, disc hemorrhage, or localized glaucomatous change is visible or suspiciously present \\
    & Overall Fundus & Overall optic nerve head appearance is consistent with glaucomatous optic neuropathy \\
    \midrule
\end{longtable}

\section{Physician preference evaluation}
\label{app:physician_pref}
For radiology (CheXpert), we conduct a blinded preference study using two medical doctors. Each physician evaluates 100 randomly selected cases. For each case, images generated by all methods are presented in randomized order without method identifiers. Physicians produce a strict ranking from best to worst based on clinical plausibility and visual realism. We aggregate rankings with a Bradley--Terry model to compute method-level preference scores and confidence intervals. Main results are shown in Figure~\ref{fig:chexpert_preference}.

Physicians were shown the following instruction: 
``For each case, you will see the target diagnosis and anonymized synthetic images generated by different methods in randomized order. Please rank the images from best to worst according to clinical plausibility, diagnostic consistency with the target label, anatomical realism, and overall image quality. Do not attempt to identify the generation method.'' 
No method names were shown during evaluation. 
The interface consisted of a randomized image grid with rank-entry fields; screenshots are omitted because the interface contained no additional task information.
The physicians were not additionally compensated for this evaluation.

\section{Automated preference-based evaluation}
\label{app:auto_pref}
In addition to metric-based evaluation, we provide an automated preference analysis using an MLLM as an independent judge. The MLLM is used only for evaluation and is not involved in training or reward computation. This analysis is intended as supporting evidence and does not replace the physician evaluation reported for CheXpert.

For each test case, the MLLM is presented with (i) the target diagnosis, (ii) the textual prompt used for synthesis, and (iii) four synthetic candidate images generated by different methods. Candidates are presented in randomized order and labeled A to D. The MLLM is instructed to produce a strict ranking of the four candidates from best to worst based on clinical plausibility, visual realism, and consistency with the provided description.

The ranking criteria are adapted to each domain.
For radiology data (CheXpert), the MLLM is instructed to act as an expert radiologist and evaluate candidates according to diagnostic accuracy (whether the image correctly exhibits the target pathology) and visual realism (anatomical structure, texture, and noise characteristics consistent with real chest X-rays).
For dermatology data (Fitzpatrick17k), the MLLM is instructed to act as an expert dermatologist and evaluate candidates based on clinical accuracy (characteristic lesion morphology, border irregularity, and color variation) and visual realism (skin texture, hair, and lighting consistency).

\begin{figure*}[!t]
  \begin{center}
    \centerline{\includegraphics[width=0.8\linewidth]{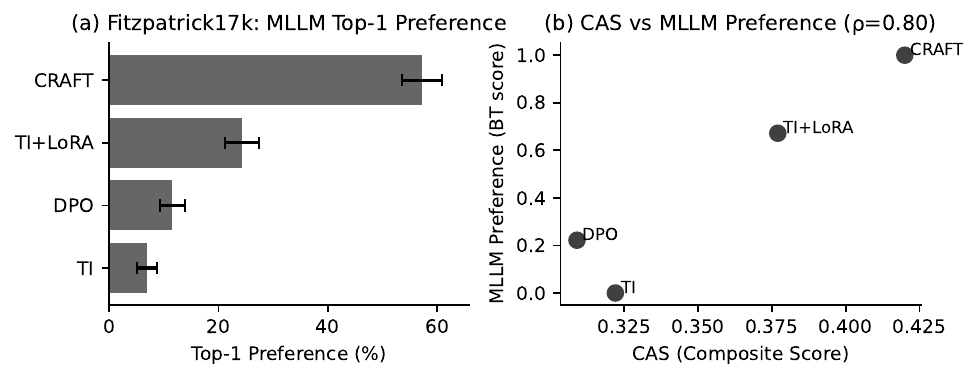}}
    \caption{
      Automated MLLM-based preference analysis on Fitzpatrick17k. (a) Top-1 preference rate with 95\% confidence intervals. (b) Correlation between the Clinical Alignment Score (CAS) and MLLM-derived Bradley--Terry preference scores across methods.
    }
    \label{app:fitz_preference}
  \end{center}
\end{figure*}

\begin{figure*}[!t]
  \begin{center}
    \centerline{\includegraphics[width=0.8\linewidth]{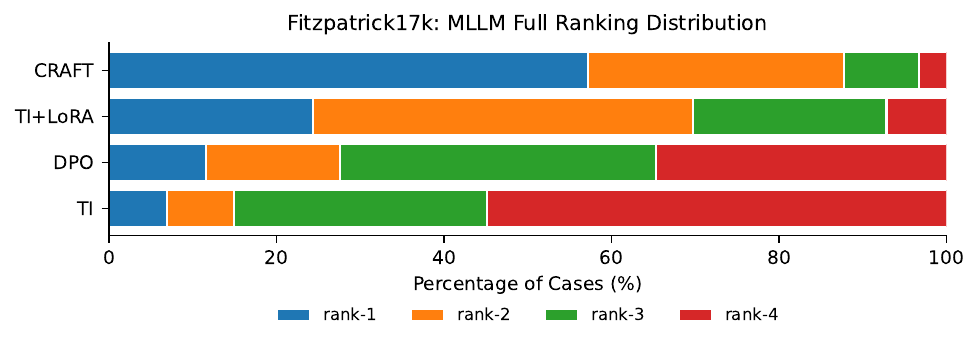}}
    \caption{
      Automated ranking distribution on Fitzpatrick17k using an MLLM judge. Each bar shows the percentage of cases in which a method is ranked at each position (Rank 1 to Rank 4).
    }
    \label{app:fitz_rank_dist}
  \end{center}
\end{figure*}

We report two complementary views of automated preference. First, we compute the top-1 selection rate, defined as the fraction of cases in which a method is ranked first by the MLLM. Second, we aggregate full rankings across cases using a BT model to obtain a continuous preference score for each method. We do not use any textual outputs from the MLLM for scoring or supervision. All quantitative analysis is based exclusively on the predicted rankings. Figure~\ref{app:fitz_preference} and Figure~\ref{app:fitz_rank_dist} summarize the resulting automated preference analysis on Fitzpatrick17k.

\begin{table}[t]
\centering
\small
\caption{Separation of model roles to mitigate evaluation leakage.}
\label{tab:role_separation}
\resizebox{\linewidth}{!}{
\begin{tabular}{lcccc}
\toprule
Component & Prompt/Checklist Gen. & Reward/Optimization & Metric Computation & Purpose \\
\midrule
MLLM (text)      & \checkmark & \texttimes & \texttimes & enriched prompts, checklists \\
MLLM (ranking)   & \texttimes & \texttimes & \checkmark & preference rankings \\
MedSigLIP (CAM)  & \texttimes & \checkmark & \texttimes & differentiable clinical reward \\
SigLIP (metrics) & \texttimes & \texttimes & \checkmark & VDC/CCS/SFS/DD/CAS metrics \\
\bottomrule
\end{tabular}}
\end{table}

\begin{table}[t]
\centering
\small
\setlength{\tabcolsep}{5pt}
\caption{Few-shot CAS under low-data adaptation across four datasets. CRAFT remains the best-performing method in both 10-shot and 20-shot regimes.}
\begin{tabular}{llcccc}
\toprule
Dataset & Shots & TI & DPO & TI+LoRA & CRAFT \\
\midrule
Fitzpatrick17k & 10-shot & 0.287 & 0.229 & 0.316 & \textbf{0.338} \\
Fitzpatrick17k & 20-shot & 0.301 & 0.242 & 0.342 & \textbf{0.350} \\
CheXpert       & 10-shot & 0.220 & 0.293 & 0.278 & \textbf{0.353} \\
CheXpert       & 20-shot & 0.237 & 0.298 & 0.340 & \textbf{0.363} \\
BreakHis       & 10-shot & 0.363 & 0.364 & 0.371 & \textbf{0.378} \\
BreakHis       & 20-shot & 0.364 & 0.368 & 0.373 & \textbf{0.381} \\
ORIGA       & 10-shot & 0.319 & 0.341 & 0.346 & \textbf{0.374} \\
ORIGA       & 20-shot & 0.376 & 0.413 & 0.415 & \textbf{0.453} \\
\bottomrule
\end{tabular}
\label{tab:fewshot_cas}
\end{table}


\section{Few-shot adaptation}
\label{app:fewshot}
To test whether CRAFT remains effective in more data-limited settings, we evaluate 10-shot and 20-shot adaptation on Fitzpatrick17k, CheXpert, BreakHis, and ORIGA. Table~\ref{tab:fewshot_cas} shows that CRAFT remains the best-performing method across all four datasets and both shot regimes.

\section{CAS--downstream utility correlation}
\label{app:cas_correlation}
To test whether CAS tracks practical downstream utility, we compute the
correlation between method-level CAS and downstream classifier performance
across five shared methods.

\begin{table}[t]
\centering
\small
\caption{Correlation between CAS and downstream classifier metrics.}
\begin{tabular}{llcccc}
\toprule
Dataset & Metric & Pearson $r$ & $p$ & Spearman $\rho$ & $p$ \\
\midrule
Fitzpatrick17k & Accuracy & 0.974 & 0.005 & 0.900 & 0.037 \\
 & Macro-F1 & 0.951 & 0.013 & 0.900 & 0.037 \\
CheXpert & Accuracy & 0.972 & 0.006 & 0.667 & 0.219 \\
 & Macro-F1 & 0.968 & 0.007 & 0.821 & 0.089 \\
\bottomrule
\end{tabular}
\label{tab:cas_corr}
\end{table}

\section{Low-alignment tail analysis}
\label{app:tail_analysis}
In addition to the main-text CDF in Figure~\ref{fig:cas_cdf}, we report full per-method tail statistics. For each dataset, the low-CAS threshold $\tau$ is defined as the 25th percentile of the real-image CAS distribution. This avoids imposing a single absolute threshold across modalities with different visual complexity and evaluator score scales.

Table~\ref{tab:tail-analysis} reports the full per-method low-CAS rates and 10th-percentile CAS values. CRAFT reduces the low-alignment region across all four datasets.

\begin{table}[t]
\centering
\small
\setlength{\tabcolsep}{4pt}
\caption{Low-alignment tail analysis. We report two complementary views of the 
lower tail of the per-image CAS distribution: the fraction of generated samples 
with CAS below $\tau$, where $\tau$ is the 25th percentile of the Real reference 
CAS distribution for each dataset, and the 10th percentile of each method's 
per-image CAS distribution. CRAFT consistently reduces the low-alignment tail 
across all four datasets, indicating fewer hallucination-like or clinically 
implausible generations under our proxy metric.}
\label{tab:tail-analysis}
\begin{tabular}{llccccc}
\toprule
Dataset & $\tau$ (Real P25) & Method & $n$ & Mean CAS $\uparrow$ & CAS $<$ $\tau$ $\downarrow$ & 10th pct. CAS $\uparrow$ \\
\midrule
\multirow{5}{*}{Fitzpatrick17k} & \multirow{5}{*}{0.310} 
 & Zero-Shot & 3100 & 0.202 & 90.0\% & 0.109 \\
 & & DPO      & 3100 & 0.331 & 64.4\% & 0.205 \\
 & & TI       & 3100 & 0.327 & 64.7\% & 0.203 \\
 & & TI+LoRA  & 3100 & 0.376 & 54.2\% & 0.230 \\
 & & CRAFT    & 3100 & \textbf{0.417} & \textbf{48.7\%} & \textbf{0.234} \\
\addlinespace
\multirow{5}{*}{CheXpert} & \multirow{5}{*}{0.314}
 & Zero-Shot & 495 & 0.205 & 73.5\% & 0.122 \\
 & & DPO      & 495 & 0.320 & 75.6\% & 0.235 \\
 & & TI       & 495 & 0.320 & 74.5\% & 0.232 \\
 & & TI+LoRA  & 495 & 0.354 & 68.9\% & 0.279 \\
 & & CRAFT    & 495 & \textbf{0.401} & \textbf{60.8\%} & \textbf{0.281} \\
\addlinespace
\multirow{5}{*}{BreakHis} & \multirow{5}{*}{0.530}
 & Zero-Shot & 1041 & 0.294 & 99.0\% & 0.183 \\
 & & DPO      & 1041 & 0.378 & 86.8\% & 0.252 \\
 & & TI       & 1041 & 0.367 & 98.3\% & 0.183 \\
 & & TI+LoRA  & 1041 & 0.402 & 89.0\% & 0.244 \\
 & & CRAFT    & 1041 & \textbf{0.417} & \textbf{52.1\%} & \textbf{0.283} \\
\addlinespace
\multirow{5}{*}{ORIGA} & \multirow{5}{*}{0.564}
 & Zero-Shot & 196 & 0.363 & 100.0\% & 0.187 \\
 & & DPO      & 196 & 0.462 & 100.0\% & 0.291 \\
 & & TI       & 196 & 0.456 & 94.9\% & 0.243 \\
 & & TI+LoRA  & 196 & 0.497 & 89.8\% & 0.294 \\
 & & CRAFT    & 196 & \textbf{0.515} & \textbf{77.0\%} & \textbf{0.306} \\
\bottomrule
\end{tabular}
\end{table}

\section{Hyperparameter sensitivity analysis}
\label{app:hyperparam}
\begin{figure}[t]
  \begin{center}
    \centerline{\includegraphics[width=\linewidth]{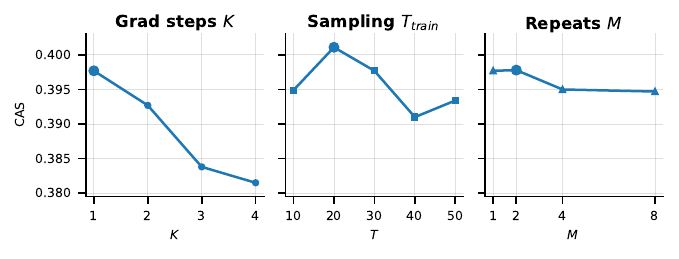}}
    \caption{
      Hyperparameter sensitivity analysis on CheXpert. Left: number of gradient backpropagation steps ($K$); Middle: number of diffusion sampling steps during training ($T_{train}$); Right: number of gradient estimation repeats ($M$). We select $K=1$, $T_{train}=20$, and $M=2$ as a balanced configuration.
    }
    \label{fig:hyperparam}
  \end{center}
  \vspace{-20pt}
\end{figure}

We investigate the sensitivity of CRAFT to three key hyperparameters on the CheXpert dataset: the number of gradient steps ($K$), training sampling steps ($T_{train}$), and gradient estimation repeats ($M$). Figure~\ref{fig:hyperparam} summarizes these results.

\textbf{Gradient Steps ($K$).} We observe that a single gradient step ($K=1$) shows the highest performance (CAS: 0.3977) with a degradation as $K$ increases (0.3815 at $K=4$). This reflects the difficulty of backpropagating through the diffusion process for multiple update steps which introduces optimization instability due to compounding variance in the gradient estimation.

\textbf{Sampling Steps ($T_{train}$).} Increasing sampling fidelity improves performance initially, peaking at $T=20$ (CAS: 0.4011). Interestingly, further increasing $T$ to 50 yields no benefit (0.3934). This shows that a coarse SDE approximation (DDPM) is sufficient to capture the correct gradient direction.

\textbf{Repeats ($M$).} Using $M=2$ repeats slightly improves performance over a single pass by reducing gradient variance (0.3978 vs. 0.3977). Further increasing $M$ provides a negligible benefit while linearly increasing training cost. Minimizing this variance too aggressively may cause the model to overfit, thereby reducing its ability to generalize.

\section{Diagnostic discriminability weight sensitivity}
\label{app:dd_weight}
Because the diagnostic discriminability term differs in scale and form from the cosine-similarity rewards, we additionally sweep its weight on CheXpert. As shown in Table~\ref{tab:dd_weight}, performance remains stable across a broad range of $\lambda_{DD}$ values, with the default setting $\lambda_{DD}=0.20$ achieving the best DD and CAS. This shows that CRAFT is not overly fragile to reasonable reward-weight choices.

\begin{table}[t]
\centering
\small
\caption{Sensitivity to the Diagnostic Discriminability (DD) reward weight on CheXpert.}
\begin{tabular}{ccc}
\toprule
$\lambda_{DD}$ & DD & CAS \\
\midrule
0.02 & 0.342 & 0.393 \\
0.10 & 0.361 & 0.398 \\
0.20 & \textbf{0.375} & \textbf{0.401} \\
0.50 & 0.365 & 0.398 \\
1.00 & 0.357 & 0.395 \\
\bottomrule
\end{tabular}
\label{tab:dd_weight}
\end{table}

\begin{figure}[t]
  \begin{center}
    \includegraphics[width=0.85\linewidth]{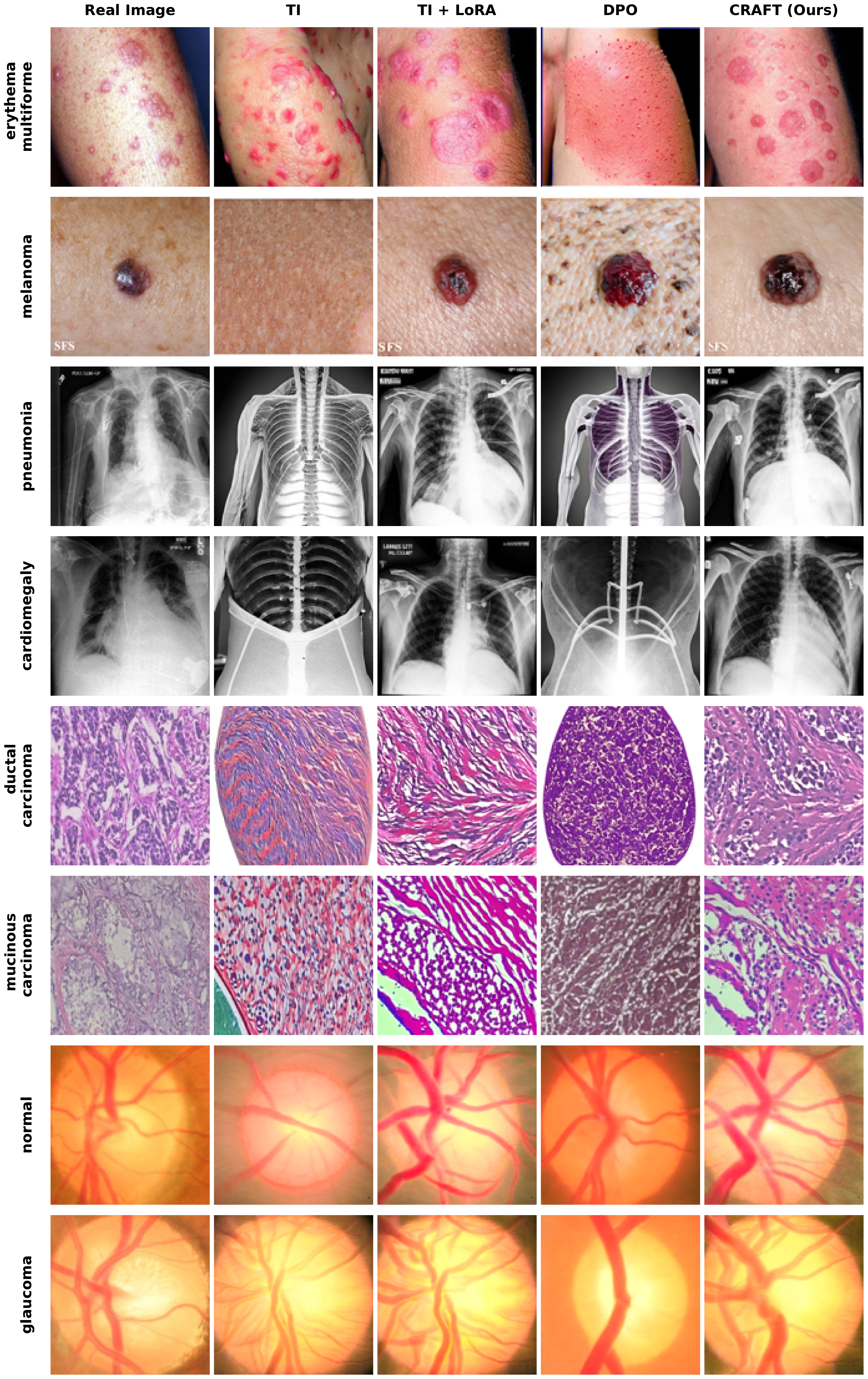}
    \caption{
        Qualitative comparison across four medical imaging domains: Fitzpatrick17k dermatology, CheXpert chest X-ray, BreakHis histopathology, and ORIGA retinal fundus imaging. Each row shows a real reference image alongside synthetic samples generated by TI~\citep{de2023medical}, TI+LoRA~\citep{wang2024majority}, DPO~\citep{wang2025doctor}, and CRAFT. CRAFT more often preserves diagnosis-consistent visual patterns while reducing ambiguous or artifact-heavy generations.
    }
    \label{fig:qualitative_main}
  \end{center}
\end{figure}

\section{Extended qualitative results}
\label{app:qualitative}
We provide additional qualitative comparisons to complement the main paper and illustrate representative synthesis behavior across methods. These examples highlight clinically relevant visual attributes such as lesion morphology, border definition, color variation, and consistency of texture in dermatology, as well as anatomical structure and pathologically appropriate opacity patterns in radiology (see Figure \ref{fig:qualitative_main}).

\begin{figure*}[!t]
  \vskip 0.2in
  \begin{center}
    \centerline{\includegraphics[width=0.8\linewidth]{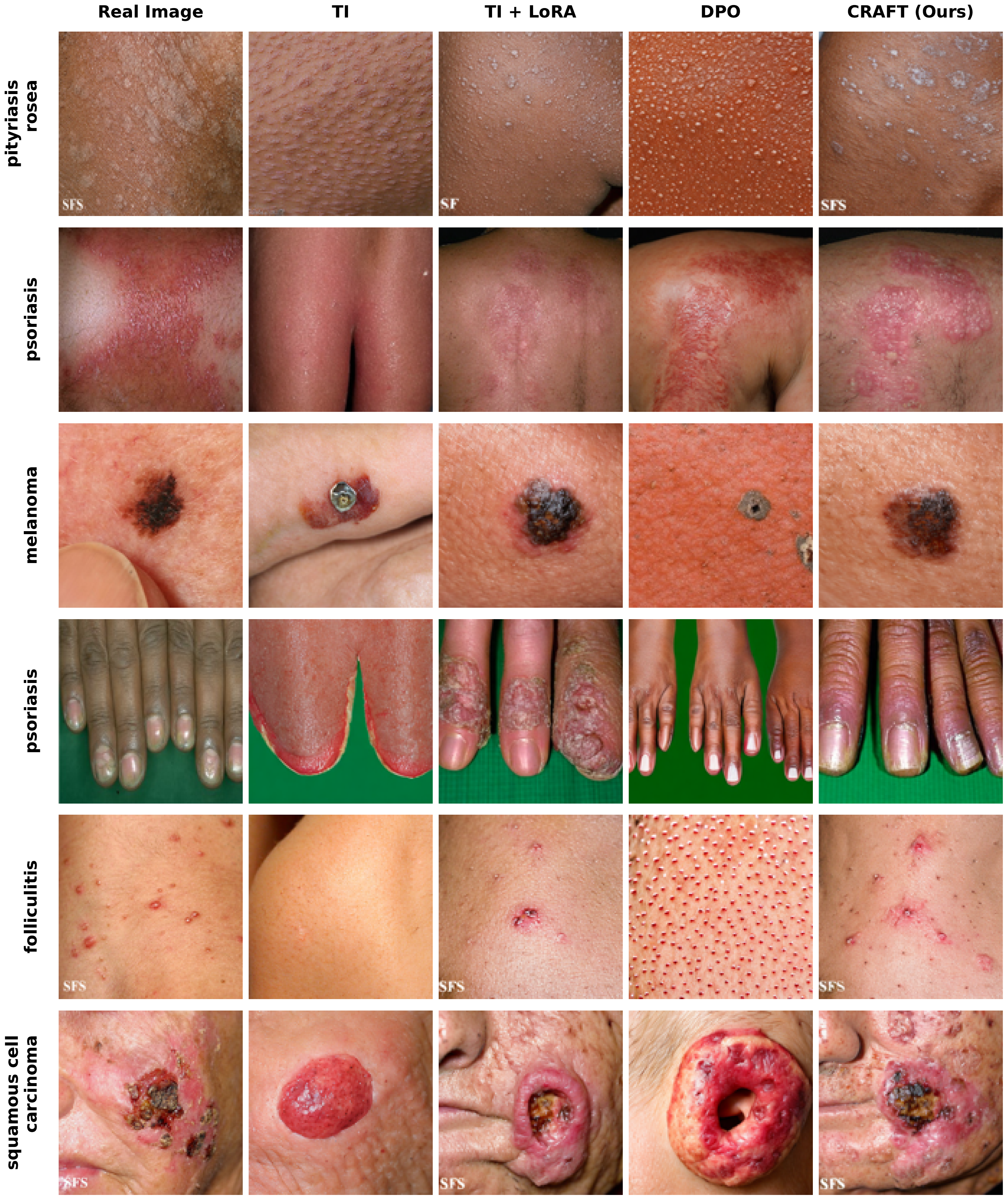}}
    \caption{
      Extended qualitative results on Fitzpatrick17k. Additional randomly selected samples comparing CRAFT against baseline methods.
    }
    \label{fig:fitz_appendix}
  \end{center}
\end{figure*}

\begin{figure*}[!t]
  \vskip 0.2in
  \begin{center}
    \centerline{\includegraphics[width=0.8\linewidth]{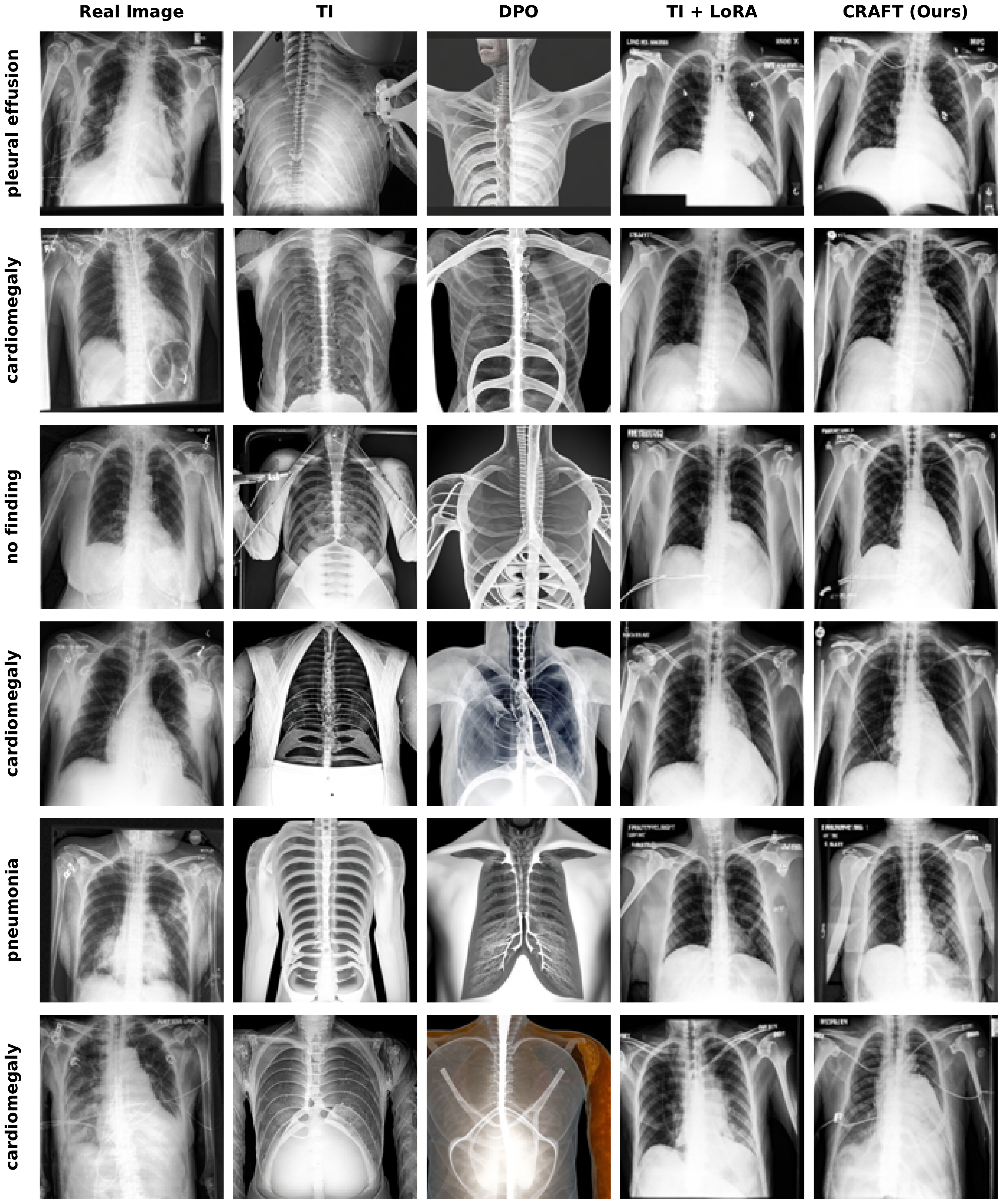}}
    \caption{
      Qualitative results on CheXpert. Randomly selected samples comparing CRAFT against baseline methods.
    }
    \label{fig:chexpert_appendix}
  \end{center}
\end{figure*}

\begin{figure*}[!t]
  \vskip 0.2in
  \begin{center}
    \centerline{\includegraphics[width=0.8\linewidth]{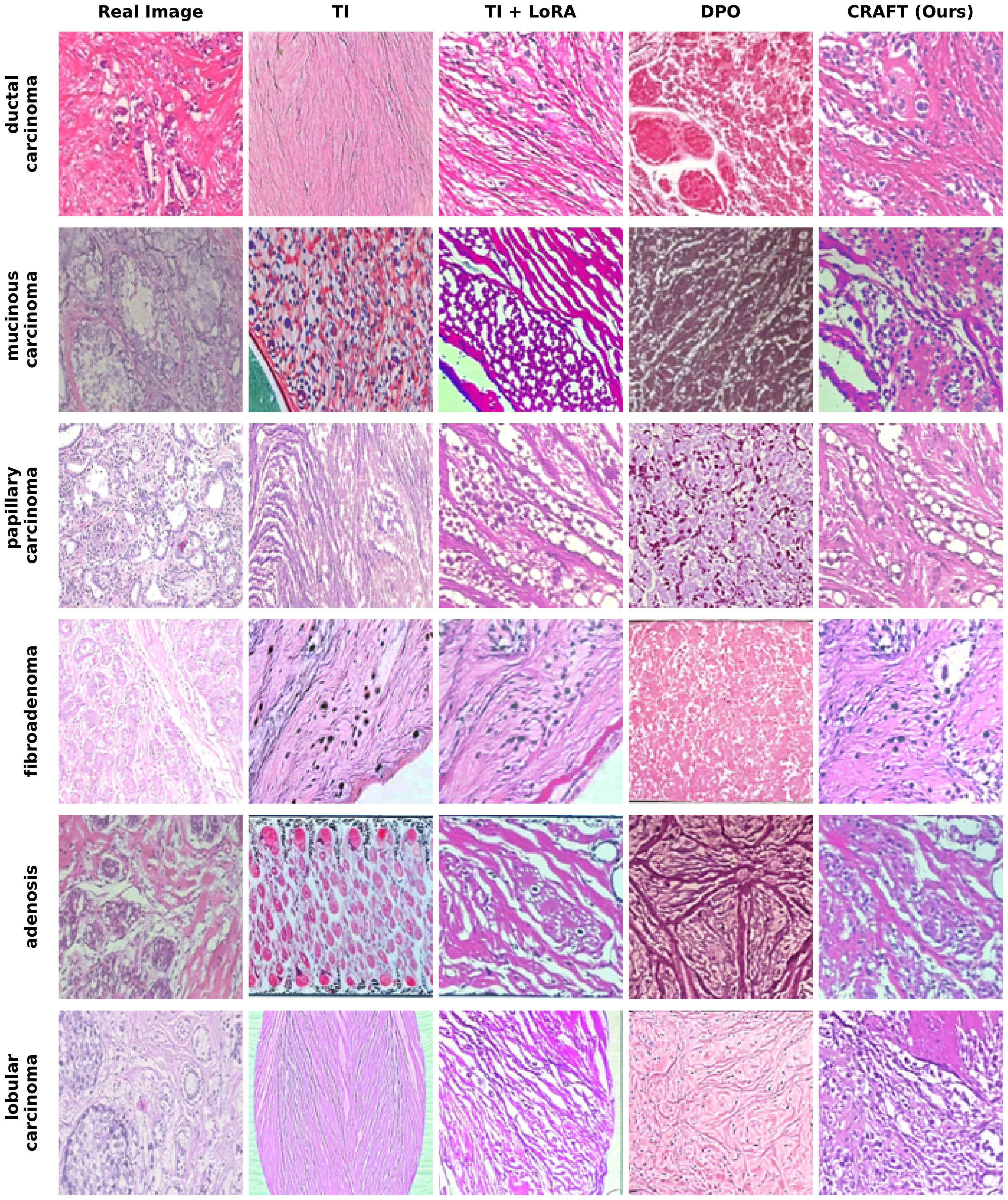}}
    \caption{
      Qualitative results on BreakHis. Randomly selected samples comparing CRAFT against baseline methods.
    }
    \label{fig:BreakHis_appendix}
  \end{center}
\end{figure*}

\begin{figure*}[!t]
  \vskip 0.2in
  \begin{center}
    \centerline{\includegraphics[width=0.8\linewidth]{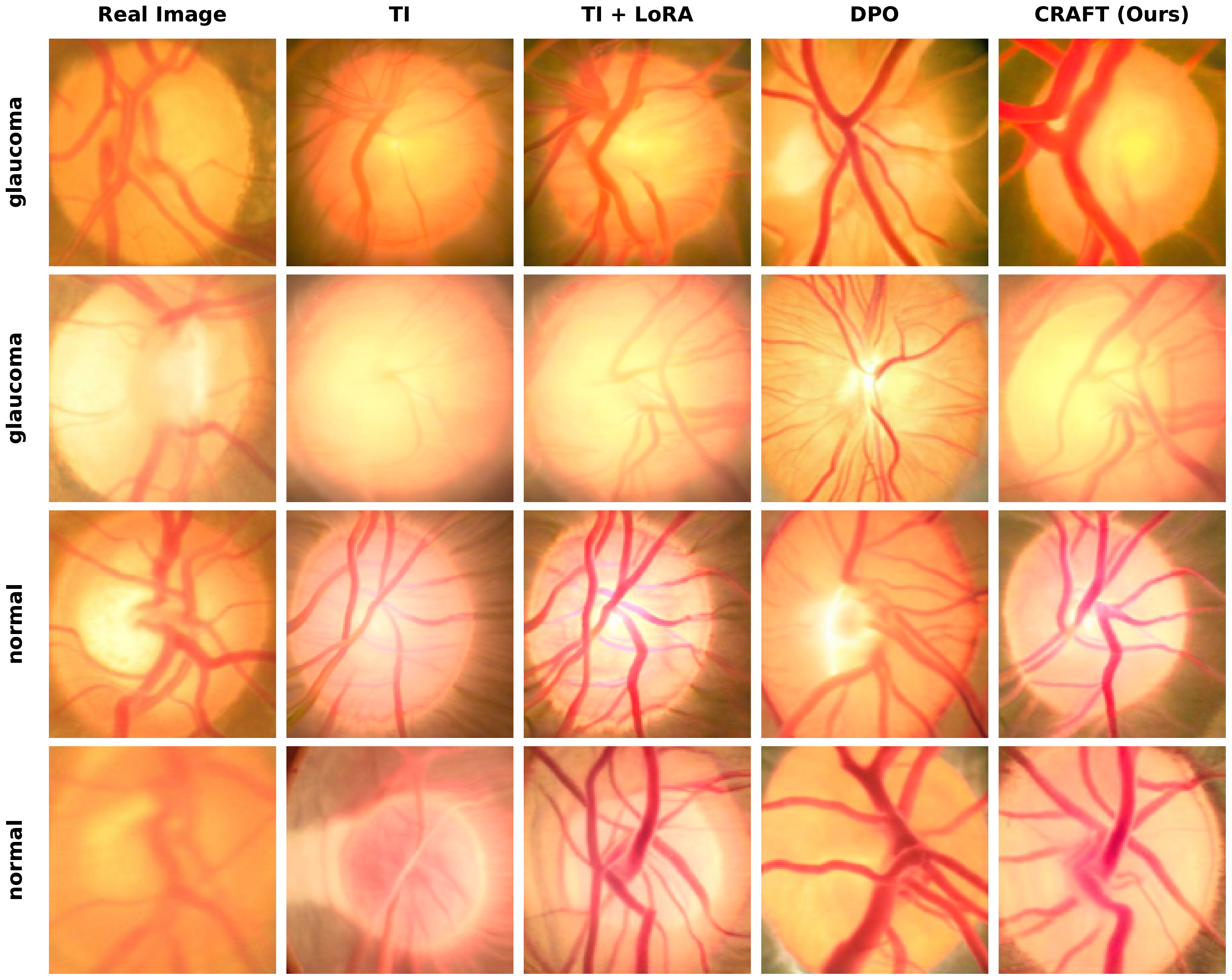}}
    \caption{
      Qualitative results on ORIGA. Randomly selected samples comparing CRAFT against baseline methods.
    }
    \label{fig:origa_appendix}
  \end{center}
\end{figure*}

Figures~\ref{fig:fitz_appendix}, \ref{fig:chexpert_appendix}, \ref{fig:BreakHis_appendix}, and \ref{fig:origa_appendix} present additional samples on Fitzpatrick17k, CheXpert, BreakHis, and ORIGA, respectively. Alongside visual realism, the comparisons emphasize whether generated images express diagnosis-consistent features and avoid ambiguous or conflicting findings that can arise when optimizing primarily for visual plausibility.

\section{Comparison with general-purpose zero-shot generators}
\label{app:zeroshot_generators}
Our main experiments compare adaptation methods on a shared Stable Diffusion v2.1 backbone to isolate the effect of clinical reward-aligned finetuning. As an additional reference, we compare against recent general-purpose zero-shot generators in Table~\ref{tab:zeroshot_refs}. Flux is the strongest zero-shot reference, but CRAFT achieves substantially higher CAS on both Fitzpatrick17k and CheXpert. This shows that larger general-purpose generators do not automatically yield stronger alignment with clinically motivated criteria in our medical prompt-conditioned setting.

\begin{table}[t]
\centering
\small
\setlength{\tabcolsep}{6pt}
\caption{CAS comparison against recent zero-shot general-purpose generators.}
\begin{tabular}{lcc}
\toprule
Method & Fitzpatrick17k CAS & CheXpert CAS \\
\midrule
Janus-Pro & 0.159 & 0.203 \\
Flux & 0.213 & 0.305 \\
SD 2.1 Zero-Shot & 0.202 & 0.205 \\
\textbf{CRAFT} & \textbf{0.417} & \textbf{0.401} \\
\bottomrule
\end{tabular}
\label{tab:zeroshot_refs}
\vspace{-12pt}
\end{table}

\section{Memorization analysis}
\label{app:memorization}
Because SFS encourages alignment between generated samples and real reference images in MedSigLIP embedding space during training, a natural concern is whether this induces memorization of training images. We evaluate this concern through two complementary tests using held-out test-prompt generations.

\paragraph{Pixel-space near-duplicate detection.}
We compute structural similarity (SSIM) and perceptual hash (pHash) distance between each generated sample and every training image, and count near-duplicates using SSIM $>$ 0.95 or pHash Hamming distance $\leq 5$. 
Table~\ref{tab:dup-detection} reports the results. No method on any dataset produces any sample with SSIM $>$ 0.95 to a training image. 
On CheXpert, CRAFT and TI+LoRA each show 2 pHash-near-duplicates out of 200 generations, indicating no CRAFT-specific memorization signal. 
CRAFT's maximum SSIM is at or below baseline methods on BreakHis and ORIGA, including the smallest dataset ORIGA with 454 training images. 
These results show no evidence that CRAFT produces verbatim or near-verbatim copies of training images.

\begin{table}[t]
\centering
\small
\caption{Pixel-space near-duplicate detection. We compute SSIM and pHash 
Hamming distance between each generated sample and every training image. ``Any'' 
counts samples flagged by either criterion. No method produces any sample with 
SSIM $>$ 0.95 to a training image on any dataset.}
\label{tab:dup-detection}
\begin{tabular}{llcccc}
\toprule
Dataset & Method & SSIM $>$ 0.95 & pHash $\leq 5$ & Any & Max SSIM \\
\midrule
\multirow{4}{*}{Fitzpatrick17k}
 & TI       & 0 & 0 & 0 & 0.901 \\
 & DPO      & 0 & 0 & 0 & 0.902 \\
 & TI+LoRA  & 0 & 0 & 0 & 0.909 \\
 & CRAFT    & 0 & 0 & 0 & 0.910 \\
\addlinespace
\multirow{4}{*}{CheXpert}
 & TI       & 0 & 0 & 0 & 0.843 \\
 & DPO      & 0 & 0 & 0 & 0.834 \\
 & TI+LoRA  & 0 & 2 & 2 & 0.872 \\
 & CRAFT    & 0 & 2 & 2 & 0.883 \\
\addlinespace
\multirow{4}{*}{BreakHis}
 & TI       & 0 & 0 & 0 & 0.469 \\
 & DPO      & 0 & 0 & 0 & 0.678 \\
 & TI+LoRA  & 0 & 0 & 0 & 0.490 \\
 & CRAFT    & 0 & 0 & 0 & \textbf{0.418} \\
\addlinespace
\multirow{4}{*}{ORIGA}
 & TI       & 0 & 0 & 0 & 0.934 \\
 & DPO      & 0 & 0 & 0 & 0.886 \\
 & TI+LoRA  & 0 & 0 & 0 & 0.929 \\
 & CRAFT    & 0 & 0 & 0 & \textbf{0.874} \\
\bottomrule
\end{tabular}
\end{table}

\paragraph{Feature-space nearest-neighbor distance in an independent encoder.}

We embed all generated, training, and held-out test images using DINOv2-base, a self-supervised vision encoder that is independent from the SigLIP family used for CAS computation and from the MedSigLIP critic used during CRAFT training. 
For each generated sample, we compute the cosine distance to its nearest training neighbor ($d_{NN}^{train}$) and nearest test neighbor ($d_{NN}^{test}$). Memorization would manifest as $d_{NN}^{train} \ll d_{NN}^{test}$, since memorized generations would collapse onto specific training images while remaining farther from unseen test images.
Table~\ref{tab:dino-nn} shows that across all four datasets, both CRAFT and TI+LoRA produce nearly symmetric distances to train and test sets, with $|\Delta| < 0.012$ for every method-dataset pair. 
On ORIGA, the smallest dataset, CRAFT's train-test gap ($-0.0037$) is smaller in magnitude than TI+LoRA's ($-0.0111$), indicating that CRAFT does not exhibit greater training-set proximity than the standard adaptation baseline.

On BreakHis and ORIGA, CRAFT generations are substantially closer to real images than TI+LoRA in DINOv2 space. Critically, this proximity is symmetric across train and test splits, indicating that CRAFT learns distributional structure of real medical images rather than memorizing specific training samples. 
Combined with the absence of SSIM near-duplicates in Table~\ref{tab:dup-detection}, these results suggest that SFS improves distributional alignment without inducing training-set copying.

\begin{table}[t]
\centering
\small
\caption{Memorization analysis via nearest-neighbor distance in DINOv2 feature 
space, an independent self-supervised encoder not used during CRAFT training or 
CAS computation. $d_{NN}^{train}$ and $d_{NN}^{test}$ are mean cosine distances 
from each generated sample to the nearest training and test image, respectively. 
Memorization would manifest as $d_{NN}^{train} \ll d_{NN}^{test}$. All methods 
show nearly symmetric distances to train and test ($|\Delta| < 0.012$), 
indicating no train-specific memorization.}
\label{tab:dino-nn}
\begin{tabular}{llccc}
\toprule
Dataset & Method & $d_{NN}^{train}$ & $d_{NN}^{test}$ & $\Delta = d_{NN}^{train} - d_{NN}^{test}$ \\
\midrule
\multirow{2}{*}{Fitzpatrick17k}
 & TI+LoRA & 0.1267 & 0.1350 & $-$0.0083 \\
 & CRAFT   & 0.1303 & 0.1370 & $-$0.0067 \\
\addlinespace
\multirow{2}{*}{CheXpert}
 & TI+LoRA & 0.0255 & 0.0255 & $\phantom{-}$0.0000 \\
 & CRAFT   & 0.0254 & 0.0259 & $-$0.0005 \\
\addlinespace
\multirow{2}{*}{BreakHis}
 & TI+LoRA & 0.1768 & 0.1757 & $\phantom{-}$0.0011 \\
 & CRAFT   & 0.0437 & 0.0439 & $-$0.0002 \\
\addlinespace
\multirow{2}{*}{ORIGA}
 & TI+LoRA & 0.1761 & 0.1872 & $-$0.0111 \\
 & CRAFT   & 0.1382 & 0.1419 & $-$0.0037 \\
\bottomrule
\end{tabular}
\end{table}

\section{Dataset label distributions}
\label{app:label_dist}
We report label distributions for the real training and test splits together with the synthetic datasets used for metric-based evaluation, to make class composition explicit and support reproducibility. For CAS-based evaluation, synthetic images are generated to match the label distribution of the held-out test split. This enables instance-level comparison without introducing class imbalance as a confound.

Tables~\ref{tab:fitz_label_distribution}, \ref{tab:chexpert_label_distribution}, \ref{tab:BreakHis_label_distribution}, and \ref{tab:origa_label_distribution} summarize the resulting distributions for Fitzpatrick17k, CheXpert, BreakHis, and ORIGA. This reporting clarifies that observed performance differences arise from image quality and clinical validity rather than unreported differences in class composition.

\begin{table}[t]
\centering
\small
\setlength{\tabcolsep}{8pt}
\caption{
Label distribution for Fitzpatrick17k. Synthetic data are generated to match the held-out test split for metric-based evaluation.
}
\begin{tabular}{lccc}
\toprule
\textbf{Condition} 
& \textbf{Train} 
& \textbf{Test} 
& \textbf{Synthetic} \\
\midrule
Psoriasis                    & 326 & 327 & 327 \\
Squamous Cell Carcinoma      & 288 & 291 & 291 \\
Lichen Planus                & 246 & 245 & 245 \\
Basal Cell Carcinoma         & 234 & 234 & 234 \\
Allergic Contact Dermatitis  & 214 & 215 & 215 \\
Lupus Erythematosus          & 205 & 205 & 205 \\
Sarcoidosis                  & 173 & 175 & 175 \\
Folliculitis                 & 171 & 171 & 171 \\
Scabies                      & 169 & 169 & 169 \\
Melanoma                     & 130 & 131 & 131 \\
Erythema Multiforme          & 118 & 118 & 118 \\
Granuloma Annulare           & 106 & 105 & 105 \\
Eczema                       & 101 & 102 & 102 \\
Pityriasis Rosea             & 96  & 97  & 97  \\
Acne                         & 92  & 91  & 91  \\
Mycosis Fungoides            & 91  & 91  & 91  \\
Actinic Keratosis            & 88  & 87  & 87  \\
Prurigo Nodularis            & 85  & 85  & 85  \\
Vitiligo                     & 80  & 83  & 83  \\
Keloid                       & 78  & 78  & 78  \\
\midrule
\textbf{Total}               & \textbf{3100} & \textbf{3100} & \textbf{3100} \\
\bottomrule
\end{tabular}
\label{tab:fitz_label_distribution}
\vspace{-10pt}
\end{table}

\begin{table}[t]
\centering
\small
\setlength{\tabcolsep}{10pt}
\caption{
Label distribution for the CheXpert subset. Synthetic data are generated to match the held-out test split for metric-based evaluation.
}
\begin{tabular}{lccc}
\toprule
\textbf{Condition} 
& \textbf{Train} 
& \textbf{Test} 
& \textbf{Synthetic} \\
\midrule
No Finding       & 125 & 124 & 124 \\
Pleural Effusion & 125 & 123 & 123 \\
Cardiomegaly     & 123 & 123 & 123 \\
Pneumonia        & 122 & 125 & 125 \\
\midrule
\textbf{Total}   & \textbf{495} & \textbf{495} & \textbf{495} \\
\bottomrule
\end{tabular}
\label{tab:chexpert_label_distribution}
\vspace{-10pt}
\end{table}

\begin{table}[t]
\centering
\small
\setlength{\tabcolsep}{8pt}
\caption{
Label distribution for the BreakHis subset. Synthetic data are generated to match the held-out test split for metric-based evaluation.
}
\begin{tabular}{lccc}
\toprule
\textbf{Condition}
& \textbf{Train}
& \textbf{Test}
& \textbf{Synthetic} \\
\midrule
Adenosis            & 56  & 57  & 57  \\
Ductal Carcinoma    & 452 & 451 & 451 \\
Fibroadenoma        & 130 & 130 & 130 \\
Lobular Carcinoma   & 85  & 85  & 85  \\
Mucinous Carcinoma  & 111 & 111 & 111 \\
Papillary Carcinoma & 71  & 71  & 71  \\
Phyllodes Tumor     & 60  & 61  & 61  \\
Tubular Adenoma     & 75  & 75  & 75  \\
\midrule
\textbf{Total}      & \textbf{1040} & \textbf{1041} & \textbf{1041} \\
\bottomrule
\end{tabular}
\label{tab:BreakHis_label_distribution}
\vspace{-10pt}
\end{table}

\begin{table}[t]
\centering
\small
\setlength{\tabcolsep}{8pt}
\caption{Label distribution for ORIGA.}
\begin{tabular}{lccc}
\toprule
\textbf{Condition} & \textbf{Train} & \textbf{Test} & \textbf{Synthetic} \\
\midrule
Non-Glaucoma & 318 & 132 & 132 \\
Glaucoma & 136 & 64 & 64 \\
\midrule
\textbf{Total} & \textbf{454} & \textbf{196} & \textbf{196} \\
\bottomrule
\end{tabular}
\label{tab:origa_label_distribution}
\end{table}

\section{Error analysis on CheXpert}
\label{app:error_analysis}

\begin{figure*}[t]
  \centering
  \includegraphics[width=0.9\linewidth]{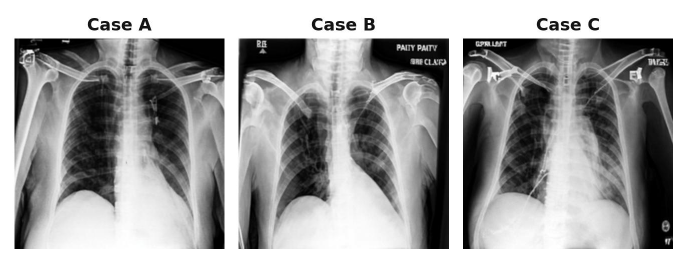}
  \caption{
  Error analysis on CheXpert. Representative CRAFT failure cases illustrating three residual failure modes: 
  (a) broadly plausible cardiomegaly with insufficient fine-grained contour evidence, 
  (b) requested pneumonia without characteristic focal consolidation, and 
  (c) visually plausible basilar opacity suggestive of pleural effusion but lacking definitive diagnostic signs.
  }
  \label{fig:chexpert_error_cases}
\end{figure*}

Although CRAFT improves overall clinical alignment, some generated chest X-rays remain clinically unreliable. Figure~\ref{fig:chexpert_error_cases} shows three representative failure modes on CheXpert.

\textbf{(1) Missing fine-grained pathology cues despite broadly plausible appearance.}
In Case A, the generated image is broadly compatible with cardiomegaly because the cardiomediastinal silhouette appears somewhat prominent. However, the radiographic detail is insufficient to support the diagnosis confidently. Adjacent left basilar/lateral opacity and partly indistinct cardiac borders limit accurate delineation of the true cardiac contour. As a result, the image appears plausible at a coarse level but lacks the fine-grained cues required for reliable diagnosis.

\textbf{(2) Morphology inconsistent with the requested diagnosis.}
In Case B, the generated image does not exhibit the typical radiographic pattern of pneumonia. There is no clear focal air-space consolidation, no convincing lobar or segmental dense opacity, and no visible air bronchograms, which are common findings supporting pneumonia on chest radiographs. Instead, the lungs remain relatively well aerated, and there is no strong silhouette sign or associated pleural effusion suggesting a focal infectious process.

\textbf{(3) Visually plausible but clinically incomplete outputs.}
In Case C, the generated image is visually plausible for pleural effusion because the lower lungs show basilar hazy opacity that can superficially resemble pleural fluid. However, the output is clinically incomplete for a confident effusion diagnosis: there is no definite meniscus sign, and the costophrenic angles are not clearly assessable. The finding may also be confused with other causes of basilar opacity or cardiac enlargement. This example illustrates that visual plausibility alone does not guarantee diagnostic completeness.

\section{Existing assets and licenses}
\label{app:assets_licenses}

Table~\ref{tab:assets_licenses} summarizes the existing datasets, models, and code assets used in this work. We cite the original sources and follow the corresponding licenses, terms of use, and data-use agreements. We do not redistribute restricted medical images or model artifacts.

\begin{table}[t]
\centering
\small
\caption{Existing assets used in this work. Dataset and model redistribution follows the original license or data-use agreement.}
\label{tab:assets_licenses}
\resizebox{\linewidth}{!}{
\begin{tabular}{llll}
\toprule
Asset & Role in paper & Source / citation & License or access terms \\
\midrule
Fitzpatrick17k & Dermatology dataset & \citet{groh2021evaluating} & Original dataset terms; not redistributed \\
CheXpert & Chest X-ray dataset & \citet{irvin2019chexpert} & Use original Stanford access terms; not redistributed \\
BreakHis & Histopathology dataset & \citet{xie2019deep} & Use original dataset license / terms \\
ORIGA & Retinal fundus dataset & \citet{zhang2010origa} & Use original dataset license / terms \\
Stable Diffusion v2.1 & Base generator & Stability AI model card & Use original model license / terms \\
MedSigLIP & Training critic & \citet{sellergren2025medgemma} & Use original model license / terms \\
SigLIP & Main evaluator & \citet{zhai2023sigmoid} & Use original model license / terms \\
MetaCLIP2 & Out-of-family evaluator & \citet{chuang2025meta} & Use original model license / terms \\
TI / LoRA / DPO baselines & Baseline methods & Cited original papers & Use original code/model licenses where applicable \\
\bottomrule
\end{tabular}
}
\end{table}

\section{Broader impact}
\label{app:broader_impact}
CRAFT may improve medical image synthesis by encouraging generated images to satisfy clinically motivated criteria rather than visual realism alone. This could support data augmentation and model development in settings where labeled medical images are scarce.

However, synthetic medical images also carry risks. They may contain subtle artifacts, reflect biases from the training datasets or foundation-model evaluators, or be overinterpreted as clinically valid evidence. 
Misuse could include presenting synthetic images as real clinical data or relying on automated clinical-alignment scores without expert review. 
We therefore frame CAS as a proxy rather than a clinical ground truth, avoid deployment claims, report failure cases, and include physician preference evaluation only as supporting evidence.


\end{document}